\documentclass[runningheads]{llncs}

\usepackage[mobile]{eccv}

\usepackage{eccvabbrv}
\usepackage{graphicx}
\usepackage{amsmath}
\usepackage{amssymb}
\usepackage{booktabs}
\usepackage{bm}
\usepackage{graphicx}
\usepackage{booktabs}
\usepackage{graphicx}
\usepackage{bbding}
\usepackage{amsmath}
\usepackage{amssymb}
\usepackage{booktabs}
\usepackage{colortbl}  
\usepackage{xcolor}
\usepackage{array}
\usepackage{amsmath,bm}
\usepackage{arydshln}

\usepackage{makecell}
\usepackage{multirow}
\usepackage{ulem}

\definecolor{hollywoodcerise}{rgb}{0.96, 0.0, 0.63}
\definecolor{lasallegreen}{rgb}{0.03, 0.47, 0.19}
\definecolor{hanpurple}{rgb}{0.32, 0.09, 0.98}
\definecolor{green(pigment)}{rgb}{0.0, 0.65, 0.31}

\definecolor{citecolor}{HTML}{229954}
\definecolor{best}{RGB}{225, 225, 225}
\definecolor{second}{RGB}{190, 205, 243}
\definecolor{aaa}{RGB}{222, 222, 230}
\definecolor{bbb}{RGB}{226, 240, 220}

\usepackage[accsupp]{axessibility}  

\usepackage[pagebackref,breaklinks,colorlinks,citecolor=eccvblue]{hyperref}

\usepackage{orcidlink}

\begin{document}

\title{Incorporating Degradation Estimation in Light Field Spatial Super-Resolution} 

\titlerunning{LF-DEST}

\author{Zeyu Xiao \quad \quad \quad Zhiwei Xiong}

\authorrunning{Xiao et al.}

\institute{University of Science and Technology of China
}

\maketitle

\begin{abstract}
Recent advancements in light field super-resolution (SR) have yielded impressive results.
In practice, however, many existing methods are limited by assuming fixed degradation models, such as bicubic downsampling, which hinders their robustness in real-world scenarios with complex degradations.
To address this limitation, we present LF-DEST, an effective blind \underline{L}ight \underline{F}ield SR method that incorporates explicit \underline{D}egradation \underline{Est}imation to handle various degradation types.
LF-DEST consists of two primary components: degradation estimation and light field restoration.
The former concurrently estimates blur kernels and noise maps from low-resolution degraded light fields, while the latter generates super-resolved light fields based on the estimated degradations.
Notably, we introduce a modulated and selective fusion module that intelligently combines degradation representations with image information, allowing for effective handling of diverse degradation types.
We conduct extensive experiments on benchmark datasets, demonstrating that LF-DEST achieves superior performance across a variety of degradation scenarios in light field SR.  


\end{abstract}

\section{Introduction}
Light field imaging allows the capture of light rays from various locations and directions~\cite{bergen1991plenoptic}, but its practical applications are limited by the spatial-angular trade-off, which results in restricted spatial resolution. This limitation hinders the potential of light field applications, such as post-capture refocusing~\cite{ng2005light}, disparity estimation~\cite{wang2015occlusion,zhang2016robust}, and occlusion removal~\cite{zhang2021removing}.
To address this issue, light field super-resolution (SR) has become a significant task focused on recovering high-resolution (HR) light fields from their low-resolution (LR) counterparts. 
With the rapid advancements in deep learning techniques, methods based on both convolutional neural networks (CNNs) and vision Transformers have shown promising results in light field SR~\cite{zhang2019residual,wang2020light,jin2020light,liu2021intra,wang2022detail,liang2023learning,Xiao_2023_cutmib}, surpassing traditional methods~\cite{rossi2017graph} with substantial improvements.

In practice, current light field SR methods prominently emphasize the development of advanced architectures to leverage both spatial and angular information effectively.
However, these methods are often designed for specific and known degradations, such as bicubic downsampling, which leaves them vulnerable to performance degradation when confronted with real-world scenes.
Consequently, there is a growing focus on blind light field SR, which tackles the SR challenge under multiple and unknown degradations.
Blind light field SR aims to address a broader range of degradation scenarios that are often encountered in real-world light field imaging conditions, such as various imaging devices (\textit{e.g.}, Lytro or RayTrix) and shot conditions (\textit{e.g.}, scene depth, focal length and illuminance)~\cite{wang2022learning}.

\begin{figure}[!t]
	\begin{center}
	\begin{minipage}{1\linewidth}
	\begin{minipage}{0.24600\linewidth}
	\vspace{-0.1mm}\centerline{\includegraphics[width=1\linewidth]{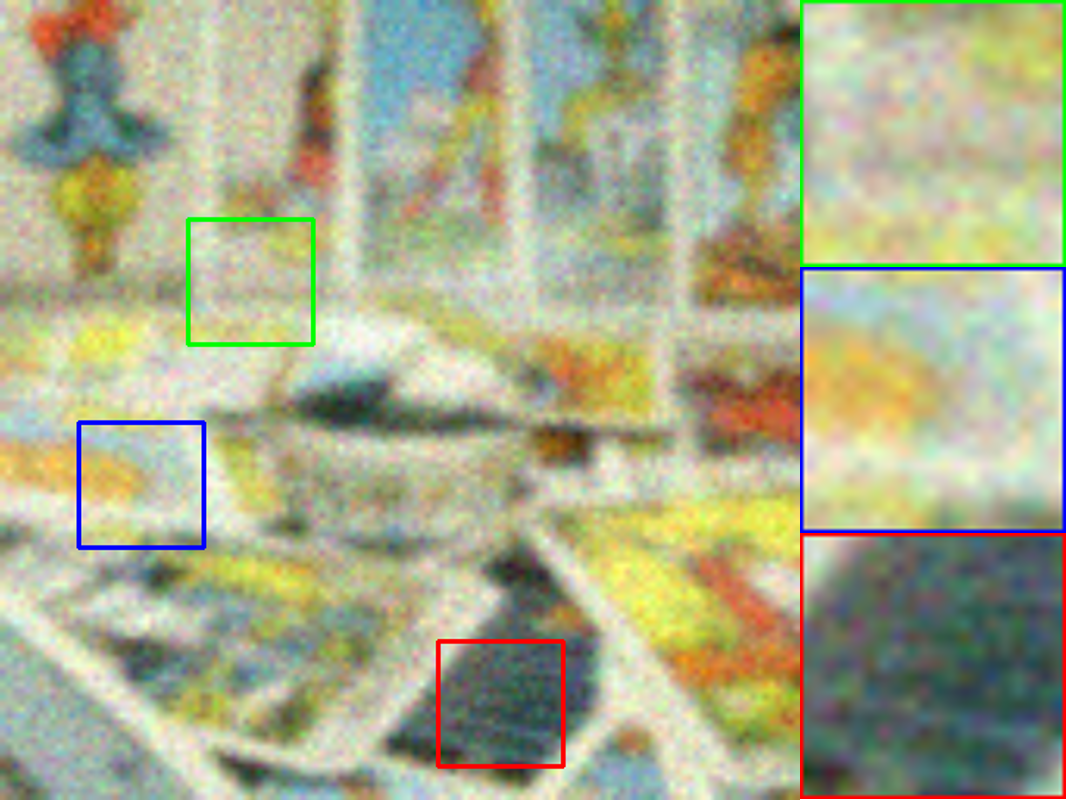}} \vfill \vspace{-0.15cm}
	\centerline{\tiny{Input}}
	\end{minipage}
	\hfill
	\hspace{-0.200cm}
	\begin{minipage}{0.24600\linewidth}
	\vspace{-0.1mm}\centerline{\includegraphics[width=1\linewidth]{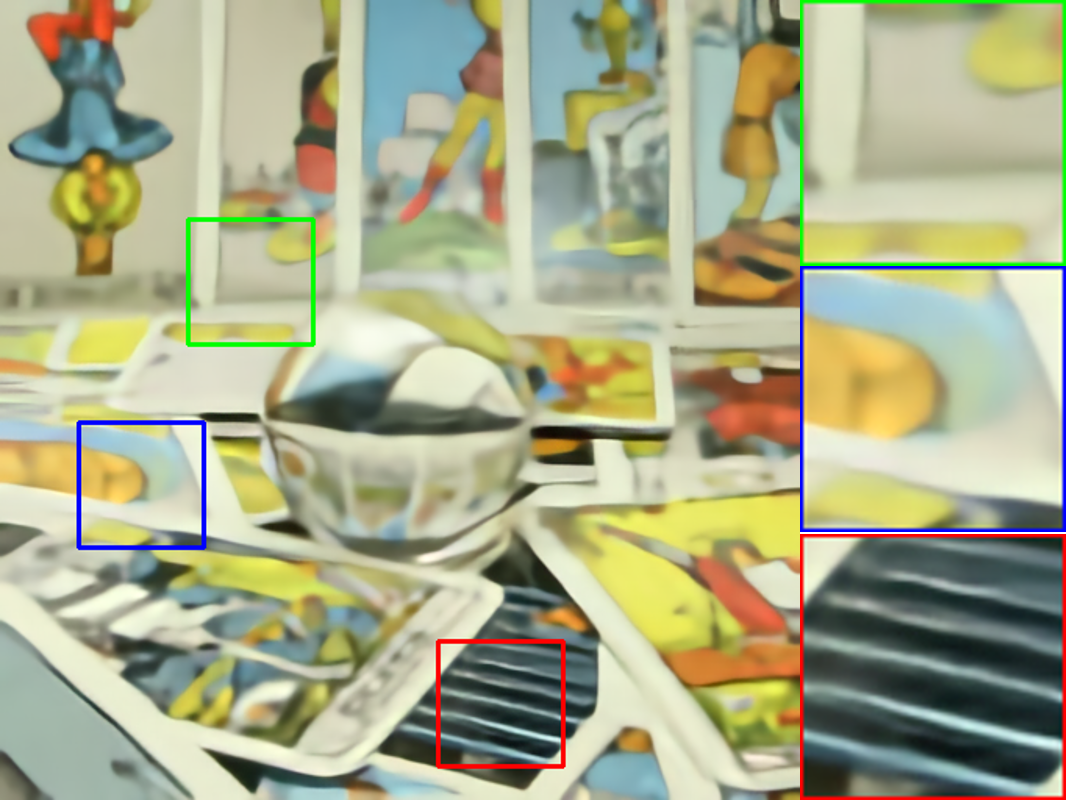}} \vfill \vspace{-0.15cm}
	\centerline{\tiny{LF-DMnet}}
	\end{minipage}
	\hfill
	\hspace{-0.200cm}
 \begin{minipage}{0.24600\linewidth}
	\vspace{-0.1mm}\centerline{\includegraphics[width=1\linewidth]{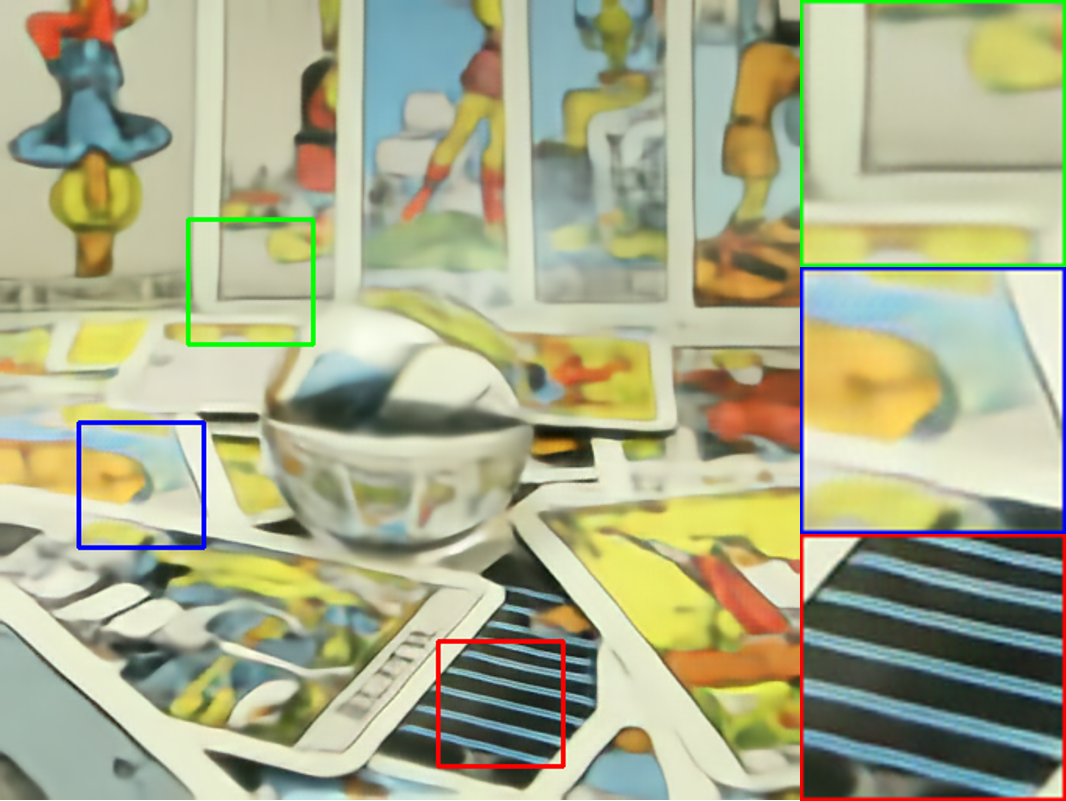}} \vfill \vspace{-0.15cm}
	\centerline{\tiny{LF-DEST}}
	\end{minipage}
	\hfill
	\hspace{-0.200cm}
	\begin{minipage}{0.24600\linewidth}
	\vspace{-0.1mm}\centerline{\includegraphics[width=1\linewidth]{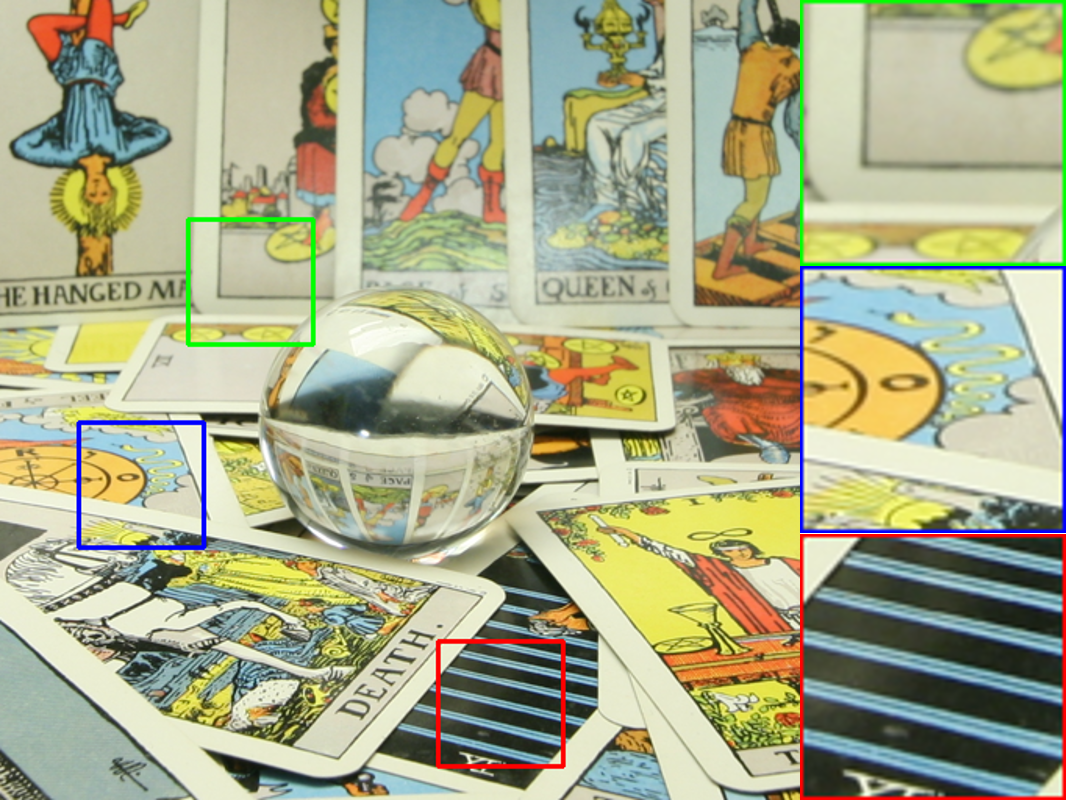}} \vfill \vspace{-0.15cm}
	\centerline{\tiny{GT}}
	\end{minipage}
	\end{minipage}
	\end{center}
	\vspace{-6mm}
	\caption{
	  \textbf{Examples of light field SR}.
	  Visual comparisons (central views) of $\times 4$ SR on a severe degraded light field scenes (kernel width=$4.5$ and noise level=$15$).
	  Existing degradation-aware light field SR method, LF-DMNet, tends to generate blurry results with obvious artifacts.
      Our proposed LF-DEST can recover faithful details from the blurry and noisy input light field.
	  }
	\label{fig:teaser}
\end{figure}

In contrast to assuming known degradations, Wang~\textit{et al.}~\cite{wang2022learning} propose an end-to-end framework called LF-DMnet, to achieve light field SR under multiple degradations (\textit{e.g.}, blur and noise).
While LF-DMnet represents a pioneering work in this direction and demonstrates promising performance, it still exhibits a notable deficiency.
LF-DMnet relies on predefined degradation parameters, including kernel width and noise level, which can limit its performance.
These fixed parameters may not adequately account for the various complex real-world degradation scenarios that affect light field images.
Consequently, the ability of LF-DMnet to generalize and effectively handle a broad range of degradation types may be compromised.
As illustrated in Figure~\ref{fig:teaser}, LF-DMnet produces sub-optimal results.

To overcome the aforementioned limitation, we propose an effective light field SR method called LF-DEST, which incorporates explicit \underline{D}egradation \underline{Est}imation.
LF-DEST is specifically designed to perform simultaneous estimation of blur kernels, noise maps, and clear latent light fields while super-resolving HR light fields.
LF-DEST comprises two key components: degradation estimation and light field restoration.
The degradation estimation process involves the prediction of blur kernels and noise maps for all sub-aperture images (SAIs) based on the light field degradation model.
This is accomplished using the side-to-center fusion block and the center-to-side fusion block, which exploit spatial-angular dependencies.
For the restoration part, LF-DEST leverages the estimated degradations to generate latent HR light fields with sharp structural details, achieved through deconvolution operations.
To ensure robustness across diverse degradation scenarios, LF-DEST incorporates modulated and selective fusion (MSF) modules. These modules adaptively combine degradation representations with LR light field information, facilitating the effective handling of various degradation types.
Furthermore, LF-DEST integrates spatial and angular information seamlessly with spatial-angular versatile (SAV) blocks. This integration enables LF-DEST to accommodate a wide range of degradation types while preserving high-quality results.
As vividly depicted in Figure~\ref{fig:teaser}, our LF-DEST outperforms LF-DMnet, delivering superior results on severely degraded light fields.
LF-DEST generates light field with finer details and fewer artifacts.

The contributions of this work are summarized as follows:
(1) LF-DEST is a novel method that addresses blind light field SR by incorporating explicit degradation estimation.
It simultaneously estimates degradation factors, \textit{i.e.}, blur kernels and noise maps, enabling accurate SR of HR light fields.
(2) We propose the MSF module, which performs adaptive fusion of degradation representations with information extracted from LR light fields.
(3) Extensive experiments demonstrate that the proposed LF-DEST achieves superior performance on various scenarios.

\section{Related Work}\label{sec:RelatedWork}

\noindent \textbf{Light field SR.}
Traditional non-learning methods depend on geometric~\cite{liang2015light,rossi2017graph} and mathematical~\cite{alain2018light} modeling of the 4D light field structure to super-resolve the reference view through projection and optimization techniques.
Deep methods now dominate light field SR due to their promising performance.
As a pioneering work along this line, \cite{yoon2017light} propose the first light field SR network LFCNN by reusing the SRCNN~\cite{dong2014learning} architecture with multiple channels.
After that, several methods have been designed to exploit across-view redundancy in the light field, either explicitly~\cite{cheng2019light,jin2020light,wang2018lfnet,zhang2019residual} or implicitly~\cite{meng2019high,wang2020spatial,yeung2018light,yuan2018light,wang2022disentangling,van2023light,van2023end}.
Transformer-based methods have recently demonstrated the effectiveness in light field SR~\cite{liang2022light,wang2022detail,liang2023learning,cong2023lfdet}.
Recently, \cite{Xiao_2023_cutmib} propose a data augmentation strategy to improve the reconstruction performance.
However, these methods primarily concentrate on advanced network design while overlooking the crucial aspect of generalization capability to multiple degradations.
In this paper, we present LF-DEST, a method to address the challenges posed by diverse and unknown degradations encountered in real-world scenarios.

\noindent \textbf{Blind image SR.}
Most blind SR approaches adopt a two-stage framework, involving kernel estimation and kernel-based HR image restoration.
For kernel estimation, KernelGAN~\cite{bell2019blind} estimates the degradation kernel by employing an InternalGAN on a single image and applies this kernel to a non-blind SR approach like ZSSR~\cite{shocher2018zero} to obtain SR results.
\cite{gu2019blind} propose applying spatial feature transform and iterative kernel correction strategy for accurate kernel estimation and SR refinement.
\cite{liang2021flow} improve the kernel estimation performance by introducing a flow-based prior.
Furthermore, \cite{tao2021spectrum} propose a spectrum-to-kernel network and demonstrate the advantages of estimating the blur kernel in the frequency domain over the spatial domain.
In the second stage, \cite{luo2020unfolding,luo2021endtoend} develop an end-to-end training deep alternating network that estimates the reduced kernel and restores the HR image.
Inspired by these methods, LF-DEST incorporates two main components, \textit{i.e.}, degradation estimation and restoration, for blind light field SR.

\noindent \textbf{Realistic light field SR.}
The difference between predefined degradations and realistic ones poses significant challenges when applying light field SR models trained with simulated data on real scenes.
Consequently, much research focus has shifted towards the task of realistic light field SR, where the degradation kernel of the input light field at test time is unknown.
Xiao~\textit{et al.}~\cite{Xiao_2023_toward} address this challenge by capturing the LytroZoom dataset, which contains well-aligned LR-HR light field pairs, and implicitly learn the SR mapping function in a data-driven manner using the OFPNet.
Wang~\textit{et al.}~\cite{wang2022learning} introduce LF-DMnet, a method designed to adapt to different degradations for light field SR when provided with corresponding degradation parameters, such as kernel width and noise level.
Cheng~\textit{et al.}~\cite{cheng2021light} propose a zero-shot learning framework to address the domain shift problem.
In this paper, we introduce LF-DEST, an effective light field SR method that incorporates explicit degradation estimation, handling diverse and unknown degradations.

 \begin{figure*}[!t]
 \centering
 \includegraphics[width=\linewidth]{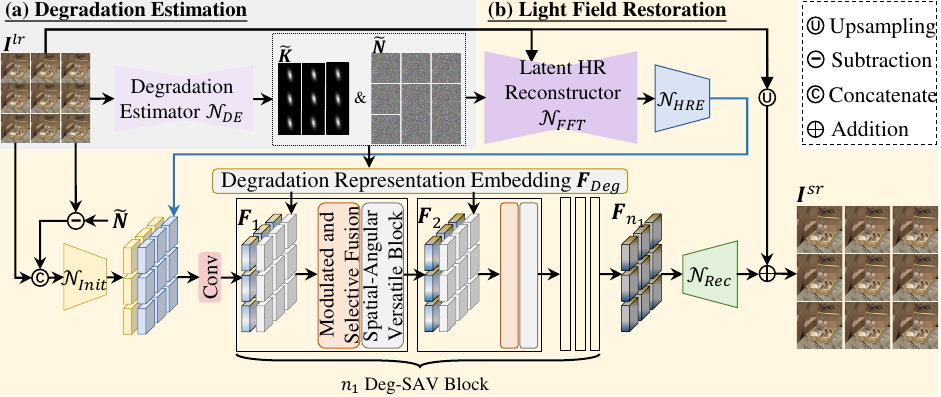}
 \vspace{-6mm}
 \caption{An overview of the proposed LF-DEST.
 LF-DEST consists of two parts: (a) degradation estimation and (b) light field restoration.
For simplicity, we use a light field with a $3 \times 3$ angular resolution as an example to illustrate LF-DEST.
 }
  \vspace{-4mm}
\label{fig:pipeline}
 \end{figure*}

\section{Preliminaries}
\subsection{Degradation Model}
Without loss of generality, we use the two-plane model \cite{levoy1996light} to parameterize a light field.
A light field can be formulated as a 4D tensor $\bm{I} \in \mathbb{R}^{U\times V \times H\times W \times 3}$, where $U$ and $V$ are angular dimensions, and $H$ and $W$ are spatial dimensions.
Each SAI is denoted as $\mathcal{I}_{u,v} \in \mathbb{R}^{H\times W \times 3}$, where $(u,v)$ is the position of the SAI.

The blind light field SR problem is formulated as follows.
Mathematically, the HR light field $\bm{I}^{hr}$ and LR light field $\bm{I}^{lr}$ are related by a degradation model
\vspace{-1mm}
 \begin{equation}\label{eq:degrad}
     \bm{I}^{lr} = (\bm{I}^{hr} \otimes \bm{K}){\downarrow}_{\alpha} + \bm{N},
 \vspace{-1mm}
\end{equation}
where $\otimes$ denotes convolution operation.
There are three main components in this degradation model, namely the blur kernel matrix $\bm{K}$, the downsampling operation ${\downarrow}_{\alpha}$ and the additive noise matrix $\bm{N}$.
For each SAI $(u,v)$, the degradation model can be represented as follows
\begin{equation}\label{eq:degrad}
     \mathcal{I}^{lr}_{u,v} = (\mathcal{I}^{hr}_{u,v} \otimes {K}_{u,v}){\downarrow}_{\alpha} + N_{u,v},
 \end{equation}
 where $K_{u,v} \in \mathbb{R}^{k \times k}$ and $N_{u,v} \in \mathbb{R}^{H\times W\times3}$ represent the blur kernel and additional noise of view $(u,v)$.
 $k$ represents the blur kernel size.

\subsection{Problem Formulation}
Given the LR light field $\bm{I}^{lr} \in \mathbb{R}^{U\times V \times H\times W \times 3}$ and the scale factor $\alpha$, our goal is to extract and fuse the complementary information between different SAIs and restore an HR light field $\bm{I}^{sr} \in \mathbb{R}^{U\times V \times \alpha H\times \alpha W \times 3}$ with rich details, which should be close to the ground-truth one $\bm{I}^{hr}$.
Following \cite{wang2022learning}, we assume that the degradation adopts isotropic Gaussian kernels and the noise $N_{u,v}$ is independent to the LR light field $\mathcal{I}^{lr}_{u,v}$.
We assume light field images exhibit isotropy across different viewing angles due to consistent environmental and camera parameter settings during capture. 
However, acknowledging the independent noise from various lenslet sensors, we ensure a comprehensive and reliable evaluation of our degradation estimation approach.
Blind light field SR can be tackled by solving the following maximum a posteriori problem
 \begin{equation}\label{eq:degrad}
\mathop{\arg\min}_{\bm{K,N},\bm{I}^{hr}} ||(\bm{I}^{lr} \! - \!(( \bm{I}^{hr} \! \otimes \! \bm{K}){\downarrow}_{\alpha}+\bm{N}))||_{2}^{2}+\phi(\bm{I}^{hr})+\psi(\bm{K},\! \bm{N}),
 \end{equation}
where $\phi(\bm{I}^{hr})$ and $\psi(\bm{K},\! \bm{N})$ are the parameterized prior regularizers, and we omit subscripts $(u,v)$ for simplicity.
Degradation prior terms are added because it is ill-posed.
We employ a deep network to implicitly express the blur prior without specifying its exact form.
Inspired by recent success of image blind SR~\cite{luo2021end}, we decompose this ill-posed problem into the following two sequential steps
\begin{equation}
\bm{K}, \bm{N} = \mathcal{N}_{DE}(\bm{I}^{lr};\theta_e),
\end{equation}
\begin{equation}
\bm{I}^{hr} = \mathcal{R}(\bm{I}^{lr},\bm{K}, \bm{N};\theta_r),
\end{equation}
where $\mathcal{N}_{DE}(\cdot)$ denotes the degradation estimator that predicts kernels and noise maps for each SAI and $\mathcal{R}(\cdot)$ denotes the light field restoration part that restores HR light field based on the LR observation and the
estimated degradations.
$\theta_e$ and $\theta_r$ are the parameters of each part, respectively.

\section{Method}
\subsection{Overview}
The proposed LF-DEST aims to recover a clean
and sharp HR light field $\bm{I}^{sr}$ given LR observation $\bm{I}^{lr}$
\begin{equation}
\bm{I}^{sr} = {\text{LF-DEST}}(\bm{I}^{lr}).
\end{equation}

Given LR observation $\bm{I}^{lr}$, we first feed $\bm{I}^{lr}$ to the degradation estimator $\mathcal{N}_{DE}$ and estimate SAI-dependent blur kernels $\tilde{\bm{K}}$ and noise maps $\tilde{\bm{N}}$ in the degradation estimation part.
In the restoration part, we generate the latent HR light field $\tilde{\bm{I}}^{sr}$ based on the deconvolution operation.
 We further extract the sharp features of $\tilde{\bm{I}}^{sr}$ using HR feature extractor $\mathcal{N}_{HRE}$, and feed the HR feature to the light field restoration part.
We then feed $\bm{I}^{lr}$ and the noise-free light field to the initial feature extractor $\mathcal{N}_{Init}$.
The extracted initial feature $\bm{F}^{Init}$ is then concatenated with $\tilde{\bm{I}}^{sr}$, and this concatenated feature is fed to building blocks consisting of MSF modules and SAV blocks.
Finally, the obtained feature with HR representations is fed into the reconstructor $\mathcal{N}_{Rec}$.
We restore the super-resolved light field $\bm{I}^{sr}$ by adding the upsampled result of $\bm{I}^{lr}$ to the output of $\mathcal{N}_{Rec}$.
Note that, LF-DEST is jointly trained in an end-to-end manner. 
The structure of LF-DEST is shown in Figure~\ref{fig:pipeline}.

\subsection{Degradation Estimation}

To handle diverse and complex real-world degradation scenarios without relying on pre-defined parameters like LF-DMnet~\cite{wang2022learning}, we introduce explicit degradation estimation in LF-DEST. This enables the network to adaptively estimate degradation factors from the input LR data, ensuring robustness across various degradation types.

\begin{figure}[!t]
 \centering
 \includegraphics[width=\linewidth]{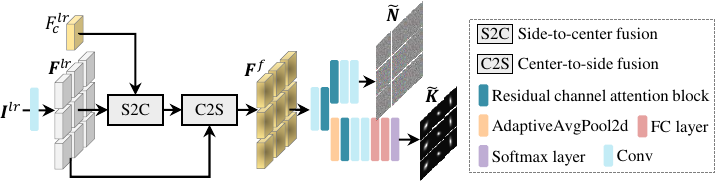}
 \vspace{-6mm}
 \caption{Details of the degradation estimator $\mathcal{N}_{DE}$.
 }
  \vspace{-4mm}
\label{fig:deg}
 \end{figure}

We introduce a Degradation Estimator $\mathcal{N}_{DE}$ to predict blur kernels and noise maps for all SAIs. 
As shown in Figure~\ref{fig:deg}, we first feed $\bm{I}^{lr}$ to a convolution layer and obtain $\bm{F}^{lr}$, which we refer to as the side-view SAI feature.
Additionally, $F_c^{lr}$ represents the center-view SAI feature.
Inspired by \cite{liu2021intra}, we sequentially feed $F_c^{lr}$ and $\bm{I}^{lr}$ into the side-to-center fusion block and the center-to-side fusion block, and obtain $\bm{F}^{f}$.
Both blocks are meticulously crafted to harness the full potential of angular and spatial correlations for accurate degradation estimation. This synergy is achieved as side-view SAIs and center-view SAI mutually complement each other, thereby enhancing the overall performance of the model.
See the supplementary document for more details.
Next, we feed $\bm{F}$ into a dual-branch structure to estimate the kernels $\tilde{\bm{K}}$ and noise maps $\tilde{\bm{N}}$.
For the branch that estimates kernels, we employ a feature extractor consists of convolution layers and residual channel attention blocks (RCABs)~\cite{zhang2018residual}, followed by the adaptive average pooling operation and a softmax operation for estimation.
As for the estimation of noise maps, we utilize cascaded convolution layers and RCABs.

To constrain the degradation estimator $\mathcal{N}_{DE}$, we introduce a self-constraint loss function as follows
\begin{equation}
\mathcal{L}_{DE} = ||(\bm{I}^{hr} \otimes \tilde{\bm{K}}) {\downarrow}_{\alpha} - \tilde{\bm{N}} - \bm{I}^{lr}||_1.
\end{equation}
In Section~\ref{sec:Experiment}, we demonstrate the effectiveness of the explicit degradation estimation.

\subsection{High-Resolution Light Field Restoration}
Thanks to the estimated degradations, LF-DEST can address a wide range of degradation types based on LR observations.
The HR light field restoration part aims at generating a sharp light field given the estimated degradations.

 As shown in Figure~\ref{fig:pipeline}, we first obtain the latent HR light field $\tilde{\bm{I}}^{sr}$ using the deconvolution operation.
 Next, we extract sharp features from $\tilde{\bm{I}}^{sr}$ through the HR feature extractor $\mathcal{N}_{HRE}$.
Simultaneously, we feed the estimated kernels and noise maps into the degradation representation embedding module, obtaining degradation representations $\bm{F}^{Deg}$.
To achieve this, we stretch the estimated kernels to the same spatial resolution as $\bm{I}^{lr}$ and then concatenate them with the noise map in the channel dimension, and the degradation representation embedding module consists of two convolutional layers.
 We further feed $\bm{I}^{lr}$ and the noise-free light field to the initial feature extractor $\mathcal{N}_{Init}$, obtaining the initial feature $\bm{F}^{Init}$.
This feature is then concatenated with $\tilde{\bm{I}}^{sr}$ and fed to a convolution layer for channel reduction, obtaining $\bm{F}_1$.
Then $\bm{F}_1$ and $\bm{F}^{Deg}$ are fed to $n_1$ building blocks consisting of MSF modules and SAV blocks, followed by $\mathcal{N}_{Rec}$ for reconstruction.

\begin{figure}[!t]
 \centering 
 \includegraphics[width=\linewidth]{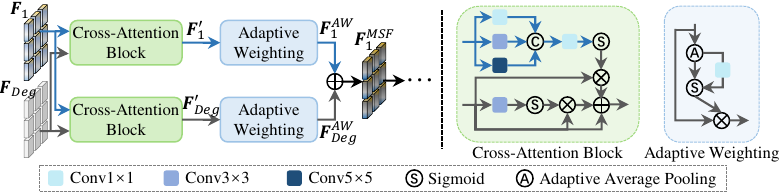}
 \vspace{-6mm}
 \caption{Details of the MSF module.
 For simplicity, we utilize $\bm{F}_1$ as an example.
 }
  \vspace{-3mm}
\label{fig:fusion}
 \end{figure}

 \noindent \textbf{Latent HR image restoration and feature extraction.}
We restore an intermediate latent HR light field $\tilde{\bm{I}}^{sr}$ with sharp structural details and then develop $\mathcal{N}_{HRE}$ to explore these restored sharp structural details for better HR light field restoration. To this end, a Fast Fourier Transformation strategy $\mathcal{N}_{FFT}$~\cite{pan2021deep} is adopted to explore the latent HR light field $\tilde{\bm{I}}^{sr}$ via
\begin{equation}
\tilde{\bm{I}}^{sr} = \mathcal{N}_{FFT}(\bm{I}^{lr}, \tilde{\bm{K}}).
\end{equation}
To better explore the sharp structural details of $\tilde{\bm{I}}^{sr}$ for final HR light field restoration, we introduce $\mathcal{N}_{HRE}$ consists of several convolution layers to obtain latent HR feature $\bm{F}^{lat}$.

 \noindent \textbf{Modulated and selective fusion module.}
The adaptive fusion of degradation representations and image features is the essence of blind light field SR.
What sets our method apart from existing approaches is the comprehensive consideration of the characteristics of both components, allowing for a flexible and adaptive fusion process.
The MSF module is pivotal in achieving this goal, enabling LF-DEST to address diverse degradation scenarios effectively.

As shown in Figure~\ref{fig:fusion}, we first feed $\bm{F}_1$ and $\bm{F}_{Deg}$ to cross-attention (CrA) blocks to fuse features of two domains (\textit{e.g.}, the image feature and the degradation representations) based on cross-domain attention, which can be denoted as
\begin{equation}
\bm{F}^{'}_1 = \mathrm{CrA}(\bm{F}_1, \bm{F}_{Deg}), \bm{F}^{'}_{Deg} = \mathrm{CrA}(\bm{F}_{Deg}, \bm{F}_1),
\end{equation}
where $\mathrm{CrA}(\cdot,\cdot)$ denotes the CrA block.
To achieve a balanced contribution between the image and degradation domains, we feed $\bm{F}^{'}_1$ and $\bm{F}^{'}_{Deg}$ to adaptive weighting (AW) layers.
This enables the method to effectively combine the information from both domains and produce more accurate results
\begin{equation}
\bm{F}^{AW}_1 = \mathrm{AW}(\bm{F}^{'}_1), \bm{F}^{AW}_{Deg} = \mathrm{AW}(\bm{F}^{'}_{Deg}),
\end{equation}
where $\mathrm{AW}(\cdot,\cdot)$ denotes the AW layer.
The output of the MSF module is finally obtained through
\begin{equation}
\bm{F}^{MSF}_1 = \bm{F}^{AW}_1+\bm{F}^{AW}_{Deg}.
\end{equation}
Details of the CrA block and the AW layeris in the supplementary document.

\begin{table*}[t]
\caption{PSNR$/$SSIM results on datasets with synthetic degradations.
The \textbf{best}, the \underline{second best} and the \uwave{third best} results are highlighted.}
\label{tab:quantitative}
		\vspace{-10mm}
		\begin{center}
			\footnotesize
			\resizebox{\textwidth}{!}{ 
                
				\begin{tabular}{|l|c|ccc|ccc|ccc|}
					\hline
					{Method} & {Kernel} &			\multicolumn{3}{c|}{HCInew}  &
					\multicolumn{3}{c|}{HCIold} & \multicolumn{3}{c|}{STFgantry}
					\tabularnewline
					\hline
                    \hline
					\multicolumn{2}{|l|}{\multirow{1}{*}{Noise level}}
					& $0$ & $15$ & $50$
					& $0$ & $15$ & $50$
					& $0$ & $15$ & $50$
					\tabularnewline
					\hline
                    \hline
					Bicubic & \multirow{9}{*}{$0$}
					& 27.71$/$0.852 & 25.90$/$0.789 & 19.53$/$0.492
					& 32.58$/$0.934 & 28.55$/$0.857 & 20.05$/$0.501
					& 26.09$/$0.845 & 24.68$/$0.789 & 19.18$/$0.516
					\tabularnewline					
					DistgSSR &
					&  \underline{31.38}$/$\textbf{0.922} & 24.88$/$0.722 & 15.59$/$0.284
					& \underline{37.56}$/$\underline{0.973} & 26.17$/$0.751 & 15.43$/$0.256
					& \underline{31.66}$/$\underline{0.953} & 24.37$/$0.754 & 15.53$/$0.319
					\tabularnewline
					LFT &
					& \textbf{31.43}$/$\underline{0.921} & 24.99$/$0.729 & 15.89$/$0.279
					& \textbf{37.63}$/$\textbf{0.974} & 26.48$/$0.765 & 15.96$/$0.258
					& \textbf{31.80}$/$\textbf{0.954} & 24.39$/$0.758 & 15.74$/$0.317
					\tabularnewline
					SRMD &
					& 29.55$/$0.886 & \uwave{27.88}$/$0.851 & \uwave{25.37}$/$\uwave{0.806}
					& 35.04$/$0.953 & \uwave{31.56}$/$\uwave{0.919} & \uwave{28.26}$/$\uwave{0.883}
					& 28.85$/$0.911 & \uwave{26.73}$/$\uwave{0.869} & \uwave{23.60}$/$\uwave{0.795}
					\tabularnewline
					DASR &
					& 29.31$/$0.886 & 27.78$/$\uwave{0.852} & 24.10$/$0.785
					& 34.54$/$0.950 & 31.45$/$\uwave{0.919} & 22.70$/$0.829
					& 26.99$/$0.897 & 26.07$/$0.866 & 21.92$/$0.768
					\tabularnewline
					BSRNet &
					& 28.42$/$0.865 & 24.98$/$0.831 & 19.32$/$0.748
					& 32.73$/$0.933 & 28.22$/$0.895 & 17.97$/$0.748
					& 26.55$/$0.880 & 22.55$/$0.829 & 17.46$/$0.714
					\tabularnewline
					Real-ESRGAN &
					& 28.05$/$0.862 & 26.99$/$0.839 & 23.65$/$0.789
					& 31.80$/$0.931 & 30.11$/$0.905 & 24.14$/$0.842
					& 24.78$/$0.871 & 24.51$/$0.850 & 19.45$/$0.754
					\tabularnewline
					LF-DMnet &
					& 30.43$/$\uwave{0.907} & \underline{29.55}$/$\underline{0.886} & \underline{28.23}$/$\underline{0.859}
					& \uwave{36.44}$/$\uwave{0.967} & \underline{34.63}$/$\underline{0.951} & \underline{32.42}$/$\underline{0.929}
					& 29.77$/$0.932 & \underline{28.62}$/$\underline{0.912} & \underline{26.99}$/$\underline{0.878}
					\tabularnewline
     LF-DEST &
					& \uwave{30.56}$/${0.905} & \textbf{30.01}$/$\textbf{0.892} & \textbf{28.79}$/$\textbf{0.870}
					& \uwave{36.44}$/$0.965 & \textbf{35.17}$/$\textbf{0.954} & \textbf{33.05}$/$\textbf{0.936}
					& \uwave{29.95}$/$\uwave{0.933} & \textbf{29.38}$/$\textbf{0.923} & \textbf{27.86}$/$\textbf{0.898}
					\tabularnewline
					\hline
                    \hline
					Bicubic & \multirow{9}{*}{$1.5$}
					& 27.02$/$0.836 & 25.42$/$0.773 & 19.41$/$0.478
					& 31.63$/$0.923 & 28.16$/$0.846 & 19.99$/$0.491
					& 25.15$/$0.821 & 24.00$/$0.764 & 18.96$/$0.493
					\tabularnewline
					DistgSSR &
					& 28.60$/$0.876 & 24.46$/$0.699 & 15.60$/$0.273
					& 33.64$/$0.949 & 25.97$/$0.739 & 15.43$/$0.251
					& 27.16$/$0.883 & 23.59$/$0.714 & 15.57$/$0.302
					\tabularnewline
					LFT &
					& 28.57$/$0.875 & 24.60$/$0.708 & 15.89$/$0.268
					& 33.62$/$0.949 & 26.25$/$0.753 & 15.96$/$0.252
					& 27.13$/$0.882 & 23.69$/$0.721 & 15.70$/$0.295
					\tabularnewline
					SRMD &
					& \uwave{29.58}$/$\uwave{0.886} & \uwave{27.39}$/$\uwave{0.840} & \uwave{25.01}$/$\uwave{0.798}
					& \uwave{35.00}$/$\uwave{0.953} & \uwave{31.02}$/$\uwave{0.912} & \uwave{27.94}$/$\uwave{0.879}
					& \uwave{28.87}$/$\uwave{0.910} & \uwave{26.05}$/$\uwave{0.851} & \uwave{23.06}$/$\uwave{0.776}
					\tabularnewline
					DASR &
					& 29.46$/$0.884 & 27.34$/$0.840 & 24.09$/$0.781
					& 34.87$/$0.952 & 30.95$/$0.911 & 23.44$/$0.831
					& 27.83$/$0.902 & 25.84$/$0.850 & 21.95$/$0.755
					\tabularnewline
					BSRNet &
					& 28.38$/$0.861 & 24.79$/$0.824 & 19.36$/$0.746
					& 32.77$/$0.932 & 28.11$/$0.892 & 18.00$/$0.749
					& 26.67$/$0.877 & 22.34$/$0.815 & 17.39$/$0.706
					\tabularnewline
					Real-ESRGAN &
					& 28.17$/$0.862 & 26.68$/$0.830 & 23.50$/$0.783
					& 32.11$/$0.932 & 29.85$/$0.900 & 24.13$/$0.840
					& 25.18$/$0.872 & 24.30$/$0.834 & 19.41$/$0.741
					\tabularnewline
					LF-DMnet &
					& \underline{30.15}$/$\textbf{0.900} & \underline{28.98}$/$\underline{0.872} & \underline{27.65}$/$\underline{0.845}
					& \underline{36.10}$/$\textbf{0.963} & \underline{33.87}$/$\underline{0.942} & \underline{31.81}$/$\underline{0.920}
					& \underline{29.47}$/$\underline{0.924} & \underline{27.91}$/$\underline{0.894} & \underline{26.25}$/$\underline{0.857}
                    \tabularnewline
				LF-DEST &
					& \textbf{30.36}$/$\underline{0.899} & \textbf{29.59}$/$\textbf{0.883} & \textbf{28.29}$/$\textbf{0.859}
					& \textbf{36.16}$/$\underline{0.962} & \textbf{34.56}$/$\textbf{0.947} & \textbf{32.49}$/$\textbf{0.928}
					& \textbf{29.69}$/$\textbf{0.927} & \textbf{28.82}$/$\textbf{0.912} & \textbf{27.20}$/$\textbf{0.884}
					\tabularnewline
					\hline
                    \hline
					Bicubic & \multirow{9}{*}{$3$}
					& 25.52$/$0.803 & 24.32$/$0.741 & 19.09$/$0.454
					& 29.59$/$0.898 & 27.12$/$0.822 & 19.82$/$0.476
					& 23.21$/$0.766 & 22.45$/$0.711 & 18.41$/$0.450
					\tabularnewline
					DistgSSR &
					& 25.79$/$0.811 & 23.30$/$0.656 & 15.47$/$0.254
					& 29.92$/$0.904 & 25.19$/$0.710 & 15.38$/$0.241
					& 23.55$/$0.780 & 21.83$/$0.639 & 15.32$/$0.265
					\tabularnewline
					LFT &
					& 25.73$/$0.810 & 23.41$/$0.665 & 15.75$/$0.248
					& 29.83$/$0.904 & 25.42$/$0.724 & 15.91$/$0.242
					& 23.47$/$0.779 & 21.89$/$0.647 & 15.43$/$0.257
					\tabularnewline
					SRMD &
					& \uwave{29.20}$/$\uwave{0.876} & \uwave{26.32}$/$\uwave{0.816} & \uwave{24.30}$/$\uwave{0.782}
					& \uwave{34.39}$/$\uwave{0.948} & \uwave{29.87}$/$\uwave{0.896} & \uwave{27.36}$/$\uwave{0.871}
					& \uwave{28.29}$/$\uwave{0.898} & \uwave{24.51}$/$\uwave{0.807} & \uwave{22.08}$/$\uwave{0.742}
					\tabularnewline	
					DASR &
					& 28.62$/$0.867 & 26.26$/$0.815 & 23.70$/$0.767
					& 33.72$/$0.942 & 29.82$/$\uwave{0.896} & 23.75$/$0.829
					& 27.71$/$0.887 & 24.48$/$\uwave{0.807} & 21.50$/$0.723
					\tabularnewline	
					BSRNet &
					& 27.60$/$0.843 & 24.13$/$0.804 & 19.33$/$0.740
					& 31.96$/$0.921 & 27.72$/$0.883 & 18.05$/$0.750
					& 26.05$/$0.849 & 21.62$/$0.776 & 17.23$/$0.691
					\tabularnewline	
					Real-ESRGAN &
					& 27.33$/$0.845 & 25.67$/$0.807 & 23.04$/$0.769
					& 31.45$/$0.919 & 29.11$/$0.889 & 23.93$/$0.835
					& 25.24$/$0.856 & 23.35$/$0.789 & 19.14$/$0.712
					\tabularnewline						
					LF-DMnet &
					& \underline{29.43}$/$\textbf{0.884} & \underline{27.76}$/$\underline{0.845} & \underline{26.54}$/$\underline{0.821}
					& \underline{35.10}$/$\textbf{0.955} & \underline{32.34}$/$\underline{0.924} & \underline{30.54}$/$\underline{0.904}
					& \underline{28.51}$/$\underline{0.904} & \underline{26.22}$/$\underline{0.853} & \underline{24.72}$/$\underline{0.814}
					\tabularnewline
                    LF-DEST &
					& \textbf{29.56}$/$\underline{0.883} & \textbf{28.52}$/$\textbf{0.861} & \textbf{27.25}$/$\textbf{0.838}
					& \textbf{35.13}$/$0.952 & \textbf{33.21}$/$\textbf{0.933} & \textbf{31.28}$/$\textbf{0.913}
					& \textbf{28.60}$/$\textbf{0.907} & \textbf{27.35}$/$\textbf{0.884} & \textbf{25.65}$/$\textbf{0.849}
					\tabularnewline
					\hline
                    \hline
					Bicubic & \multirow{9}{*}{$4.5$}
					& 24.36$/$0.779 & 23.41$/$0.718 & 18.79$/$0.438
					& 28.05$/$0.879 & 26.19$/$0.803 & 19.63$/$0.465
					& 21.80$/$0.725 & 21.26$/$0.672 & 17.90$/$0.420
					\tabularnewline
					DistgSSR &
					& 24.38$/$0.781 & 22.48$/$0.631 & 15.33$/$0.242
					& 28.08$/$0.880 & 24.50$/$0.690 & 15.31$/$0.235
					& 21.83$/$0.728 & 20.67$/$0.595 & 15.04$/$0.243
					\tabularnewline
					LFT &
					& 24.39$/$0.781 & 22.57$/$0.640 & 15.61$/$0.237
					& 28.08$/$0.880 & 24.71$/$0.705 & 15.84$/$0.235
					& 21.84$/$0.728 & 20.72$/$0.602 & 15.14$/$0.233
					\tabularnewline
					SRMD &
					& \uwave{26.32}$/$\uwave{0.818} & \uwave{25.09}$/$\uwave{0.792} & \uwave{23.65}$/$\uwave{0.769}
					& \uwave{30.62}$/$\uwave{0.908} & \uwave{28.61}$/$\uwave{0.882} & \uwave{26.66}$/$\uwave{0.864}
					& \uwave{24.34}$/$\uwave{0.780} & \uwave{22.80}$/$\uwave{0.753} & \uwave{21.28}$/$\uwave{0.716}
					\tabularnewline	
					DASR &
					& 25.34$/$0.799 & 24.89$/$0.788 & 23.11$/$0.755
					& 29.33$/$0.895 & 28.39$/$0.880 & 23.94$/$0.827
					& 22.99$/$0.761 & 22.65$/$0.749 & 20.76$/$0.697
					\tabularnewline	
					BSRNet &
					& 26.31$/$0.816 & 23.40$/$0.784 & 19.26$/$0.734
					& 30.35$/$0.902 & 27.10$/$0.874 & 18.03$/$0.749
					& 24.23$/$0.795 & 20.83$/$0.738 & 17.06$/$0.680
					\tabularnewline	
					Real-ESRGAN &
					& 26.28$/$0.816 & 24.69$/$0.787 & 22.55$/$0.758
					& 30.04$/$0.900 & 28.08$/$0.878 & 23.66$/$0.830
					& 23.97$/$0.810 & 22.17$/$0.743 & 18.90$/$0.693
					\tabularnewline						
					LF-DMnet &
					& \textbf{28.00}$/$\textbf{0.854} & \underline{26.55}$/$\underline{0.820} & \underline{25.58}$/$\underline{0.801}
					& \textbf{33.39}$/$\textbf{0.937} & \underline{30.88}$/$\underline{0.906} & \underline{29.45}$/$\underline{0.890}
					& \textbf{26.59}$/$\underline{0.860} & \underline{24.64}$/$\underline{0.808} & \underline{23.30}$/$\underline{0.771}
                    \tabularnewline						
					LF-DEST &
					& \underline{27.85}$/$\underline{0.848} & \textbf{27.11}$/$\textbf{0.832} & \textbf{25.93}$/$\textbf{0.810}
					& \underline{32.89}$/$\underline{0.931} & \textbf{31.56}$/$\textbf{0.914} & \textbf{29.96}$/$\textbf{0.898}
					& \underline{26.41}$/$\textbf{0.861} & \textbf{25.41}$/$\textbf{0.837} & \textbf{23.81}$/$\textbf{0.795}
					\tabularnewline
					\hline						
			\end{tabular}}
		\end{center}
		\vspace{-5mm}
	\end{table*}

 \noindent \textbf{Spatial-angular
versatile block.}
The SAV block is inspired by SAV-conv~\cite{cheng2022spatial}, which combines spatial-angular separable convolution (SAS-conv) and spatial-angular correlated convolution (SAC-conv) to comprehensively exploit geometric information in a light field at both local and global scales.
Compared to SAV-conv, our SAV block incorporates a channel attention operation, enabling adaptive extraction of essential information for light field SR.
See the supplementary document for more details.

We utilize the $L_1$ loss to train the method via
\begin{equation}
\mathcal{L}_{rec} = ||\bm{I}^{sr}-\bm{I}^{hr}||_1.
\end{equation}

\section{Experiments}\label{sec:Experiment}
\subsection{Experimantal Settings}

\noindent \textbf{Datasets.}
We follow \cite{wang2022learning} and utilize three public light field datasets, namely HCInew \cite{HCInew}, HCIold \cite{HCIold}, and STFgantry \cite{STFgantry}, for training and testing.
The training-testing split remains consistent with previous works~\cite{wang2020spatial, wang2020light, liang2023learning}.

\noindent \textbf{Training details.}
During the training phase, we crop HR SAIs into patches of size $152\times152$ with a stride of 32.
Using the degradation model, we synthesize LR SAI patches of size $38\times38$ with $\alpha=4$.
Following \cite{gu2019blind} and \cite{wang2022learning}, we set the window size of the isotropic Gaussian kernel to $21\times21$ and sample the kernel width and noise level from the range $[0, 4]$ and $[0, 75]$, respectively.
To avoid boundary effects caused by Gaussian filtering, we use the central $128\times128$ region of the HR patches and their corresponding $32\times32$ LR patches for training.
To augment the training data, we perform flipping, rotation, and RGB channel shuffling.
LF-DEST is trained and optimized using Adam with $\beta_1=0.9$, $\beta_2=0.999$ and a batch size of 8.
$n_1$ is set to 10.
LF-DEST is implemented in PyTorch with one NVIDIA 3090 GPU.
We first train the degradation estimator $\mathcal{N}_{DE}$ from scratch by $3\times 10^4$ iterations, with the learning rate set to $2\times10^{-4}$ and then jointly train the whole LF-DEST.
The learning rate for whole training is initially set to $2\times10^{-4}$ and decreased by a factor of 0.5 for every $3\times 10^4$ iterations.
The training is stopped after $1\times 10^5$ iterations.

\noindent \textbf{Inference details.}
Following~\cite{wang2022learning,gu2019blind}, we only consider $4 \times$ light field SR.
We evaluate our proposed LF-DEST on several datasets, including HCInew \cite{HCInew}, HCIold \cite{HCIold}, and STFgantry \cite{STFgantry}.
We use PSNR and SSIM calculated on RGB channel images as quantitative metrics~\cite{wang2022learning}.
To assess the generalization capability of our LF-DEST to real-world degradations, we employ the LytroZoom dataset, the real-world light field SR dataset~\cite{Xiao_2023_toward}.

\begin{figure*}[!t]
	\begin{center}
	\begin{minipage}{\linewidth}
	\begin{minipage}{0.26800\linewidth}
	\centerline{\includegraphics[width=1\linewidth, ]{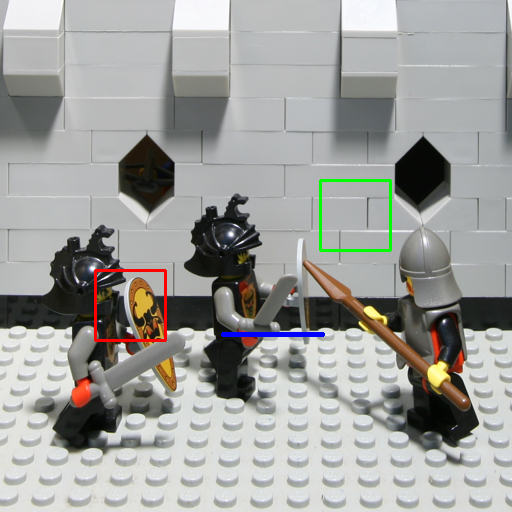}} \vfill \vspace{-0.15cm}
	\centerline{\tiny{Stanford\_Gantry\_1}}
	\end{minipage}
	\hfill
	\hspace{-0.200cm}
	\begin{minipage}{0.11900\linewidth}
	\centerline{\includegraphics[width=1\linewidth]{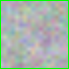}} \vspace{-0.15mm}
	\centerline{\includegraphics[width=1\linewidth]{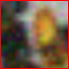}} \vfill  \vspace{0.25mm}
	\centerline{\includegraphics[width=1\linewidth]{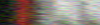}} \vfill  \vspace{-0.15cm}
	\centerline{\tiny{Input}}
	\end{minipage}
 \hfill
	\hspace{-0.200cm}
	\begin{minipage}{0.11900\linewidth}
	\centerline{\includegraphics[width=1\linewidth]{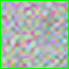}} \vspace{-0.15mm}
	\centerline{\includegraphics[width=1\linewidth]{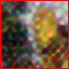}} \vfill \vspace{0.25mm}
	\centerline{\includegraphics[width=1\linewidth]{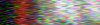}} \vfill \vspace{-0.15cm}
	\centerline{\tiny{LFT}}
	\end{minipage}
	\hfill
	\hspace{-0.200cm}
	\begin{minipage}{0.11900\linewidth}
	\centerline{\includegraphics[width=1\linewidth]{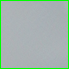}} \vspace{-0.15mm}
	\centerline{\includegraphics[width=1\linewidth]{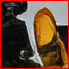}} \vfill  \vspace{0.25mm}
	\centerline{\includegraphics[width=1\linewidth]{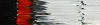}} \vfill \vspace{-0.15cm}
	\centerline{\tiny{Real-ESRGAN}}
	\end{minipage}
	\hfill
	\hspace{-0.200cm}
	\begin{minipage}{0.11900\linewidth}
	\centerline{\includegraphics[width=1\linewidth]{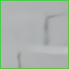}} \vspace{-0.15mm}
	\centerline{\includegraphics[width=1\linewidth]{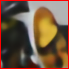}} \vfill  \vspace{0.25mm}
	\centerline{\includegraphics[width=1\linewidth]{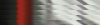}} \vfill \vspace{-0.15cm}
	\centerline{\tiny{LF-DMnet}}
	\end{minipage}
	\hfill
	\hspace{-0.200cm}
	\begin{minipage}{0.11900\linewidth}
	\centerline{\includegraphics[width=1\linewidth]{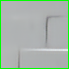}} \vspace{-0.15mm}
	\centerline{\includegraphics[width=1\linewidth]{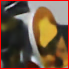}} \vfill  \vspace{0.25mm}
	\centerline{\includegraphics[width=1\linewidth]{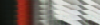}} \vfill \vspace{-0.15cm}
	\centerline{\tiny{LF-DEST}}
	\end{minipage}
	\hfill
	\hspace{-0.200cm}
	\begin{minipage}{0.11900\linewidth}
	\centerline{\includegraphics[width=1\linewidth]{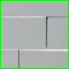}} \vspace{-0.15mm}
	\centerline{\includegraphics[width=1\linewidth]{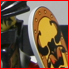}} \vfill  \vspace{0.25mm}
	\centerline{\includegraphics[width=1\linewidth]{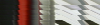}} \vfill \vspace{-0.15cm}
	\centerline{\tiny{GT}}
	\end{minipage}
	\vfill
 	\begin{minipage}{0.26800\linewidth}
	\centerline{\includegraphics[width=1\linewidth, ]{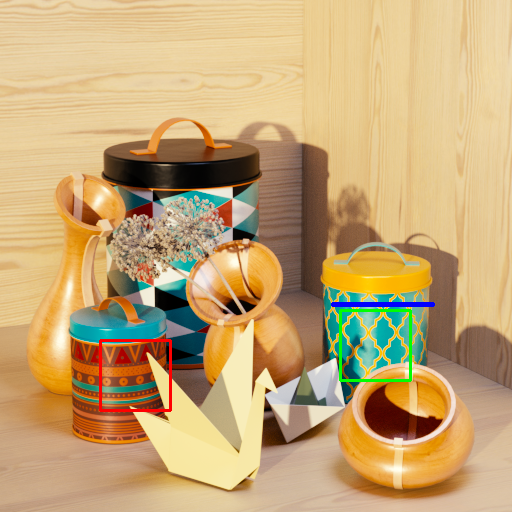}} \vfill \vspace{-0.15cm}
	\centerline{\tiny{HCI\_new\_2}}
	\end{minipage}
	\hfill
	\hspace{-0.200cm}
	\begin{minipage}{0.11900\linewidth}
	\centerline{\includegraphics[width=1\linewidth]{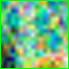}} \vspace{-0.15mm}
	\centerline{\includegraphics[width=1\linewidth]{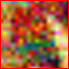}} \vfill  \vspace{0.25mm}
	\centerline{\includegraphics[width=1\linewidth]{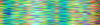}} \vfill \vspace{-0.15cm}
	\centerline{\tiny{Input}}
	\end{minipage}
 \hfill
	\hspace{-0.200cm}
	\begin{minipage}{0.11900\linewidth}
	\centerline{\includegraphics[width=1\linewidth]{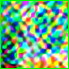}} \vspace{-0.15mm}
	\centerline{\includegraphics[width=1\linewidth]{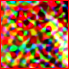}} \vfill \vspace{0.25mm}
	\centerline{\includegraphics[width=1\linewidth]{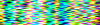}} \vfill \vspace{-0.15cm}
	\centerline{\tiny{LFT}}
	\end{minipage}
	\hfill
	\hspace{-0.200cm}
	\begin{minipage}{0.11900\linewidth}
	\centerline{\includegraphics[width=1\linewidth]{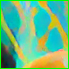}} \vspace{-0.15mm}
	\centerline{\includegraphics[width=1\linewidth]{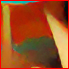}} \vfill  \vspace{0.25mm}
	\centerline{\includegraphics[width=1\linewidth]{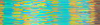}} \vfill \vspace{-0.15cm}
	\centerline{\tiny{Real-ESRGAN}}
	\end{minipage}
	\hfill
	\hspace{-0.200cm}
	\begin{minipage}{0.11900\linewidth}
	\centerline{\includegraphics[width=1\linewidth]{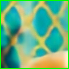}} \vspace{-0.15mm}
	\centerline{\includegraphics[width=1\linewidth]{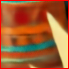}} \vfill  \vspace{0.25mm}
	\centerline{\includegraphics[width=1\linewidth]{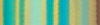}} \vfill \vspace{-0.15cm}
	\centerline{\tiny{LF-DMnet}}
	\end{minipage}
	\hfill
	\hspace{-0.200cm}
	\begin{minipage}{0.11900\linewidth}
	\centerline{\includegraphics[width=1\linewidth]{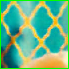}} \vspace{-0.15mm}
	\centerline{\includegraphics[width=1\linewidth]{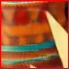}} \vfill  \vspace{0.25mm}
	\centerline{\includegraphics[width=1\linewidth]{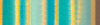}} \vfill \vspace{-0.15cm}
	\centerline{\tiny{LF-DEST}}
	\end{minipage}
	\hfill
	\hspace{-0.200cm}
	\begin{minipage}{0.11900\linewidth}
	\centerline{\includegraphics[width=1\linewidth]{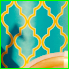}} \vspace{-0.15mm}
	\centerline{\includegraphics[width=1\linewidth]{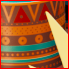}} \vfill  \vspace{0.25mm}
	\centerline{\includegraphics[width=1\linewidth]{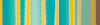}} \vfill \vspace{-0.15cm}
	\centerline{\tiny{GT}}
	\end{minipage}
	\end{minipage}
	\end{center}
	\vspace{-6mm}
	\caption{
	 Visual results achieved by different methods on synthetically degraded light field (top: kernel width=3, noise level=15 and bottom: kernel width=1.5, noise level=50) for $\times 4$ SR. The super-resolved center view images and EPIs are shown.
	  }
	\label{fig:compare1}
\end{figure*}

\begin{figure*}[!t]
	\begin{center}
	\begin{minipage}{\linewidth}
	\begin{minipage}{0.26800\linewidth}
	\centerline{\includegraphics[width=1\linewidth,height=1\linewidth]{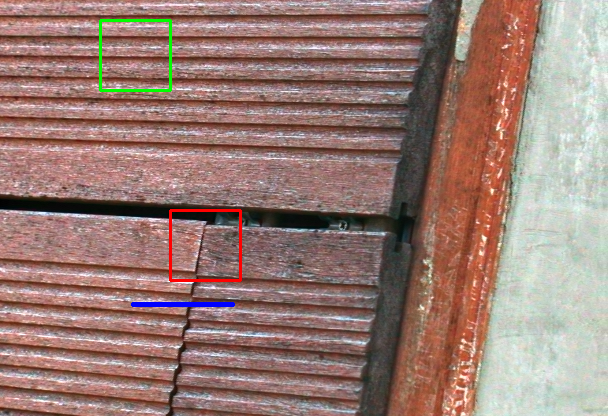}} \vfill \vspace{-0.15cm}
	\centerline{\tiny{LytroZoom-O\_Chair1}}
	\end{minipage}
	\hfill
	\hspace{-0.200cm}
	\vspace{0.15cm}\begin{minipage}{0.11900\linewidth}
	\centerline{\includegraphics[width=1\linewidth]{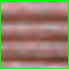}} \vspace{-0.15mm}
	\centerline{\includegraphics[width=1\linewidth]{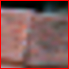}} \vfill  \vspace{0.25mm}
	\centerline{\includegraphics[width=1\linewidth]{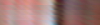}} \vfill \vspace{-0.15cm}
	\centerline{\tiny{Input}}
	\end{minipage}
 \hfill  
	\hspace{-0.200cm}
	\begin{minipage}{0.11900\linewidth}
	\centerline{\includegraphics[width=1\linewidth]{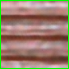}} \vspace{-0.15mm}
	\centerline{\includegraphics[width=1\linewidth]{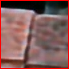}} \vfill  \vspace{0.25mm}
	\centerline{\includegraphics[width=1\linewidth]{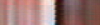}} \vfill \vspace{-0.15cm}
	\centerline{\tiny{LFT}}
	\end{minipage}
	\hfill
	\hspace{-0.200cm}
	\begin{minipage}{0.11900\linewidth}
	\centerline{\includegraphics[width=1\linewidth]{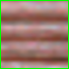}} \vspace{-0.15mm}
	\centerline{\includegraphics[width=1\linewidth]{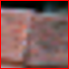}} \vfill  \vspace{0.25mm}
	\centerline{\includegraphics[width=1\linewidth]{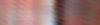}} \vfill \vspace{-0.15cm}
	\centerline{\tiny{Real-ESRGAN}}
	\end{minipage}
	\hfill
	\hspace{-0.200cm}
	\begin{minipage}{0.11900\linewidth}
	\centerline{\includegraphics[width=1\linewidth]{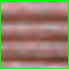}} \vspace{-0.15mm}
	\centerline{\includegraphics[width=1\linewidth]{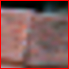}} \vfill  \vspace{0.25mm}
	\centerline{\includegraphics[width=1\linewidth]{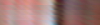}} \vfill \vspace{-0.15cm}
	\centerline{\tiny{LF-DMnet}}
	\end{minipage}
	\hfill
	\hspace{-0.200cm}
	\begin{minipage}{0.11900\linewidth}
	\centerline{\includegraphics[width=1\linewidth]{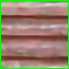}} \vspace{-0.15mm}
	\centerline{\includegraphics[width=1\linewidth]{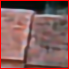}} \vfill  \vspace{0.25mm}
	\centerline{\includegraphics[width=1\linewidth]{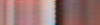}} \vfill \vspace{-0.15cm}
	\centerline{\tiny{LF-DEST}}
	\end{minipage}
	\hfill
	\hspace{-0.200cm}
	\begin{minipage}{0.11900\linewidth}
	\centerline{\includegraphics[width=1\linewidth]{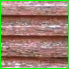}} \vspace{-0.15mm}
	\centerline{\includegraphics[width=1\linewidth]{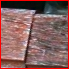}} \vfill  \vspace{0.25mm}
	\centerline{\includegraphics[width=1\linewidth]{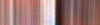}} \vfill \vspace{-0.15cm}
	\centerline{\tiny{GT}}
	\end{minipage}
	\end{minipage}
	\end{center}
	\vspace{-7mm}
	\caption{
	 Visual results achieved by different methods on real-world degraded light field for $\times 4$ SR.
  The super-resolved center view images and EPIs are shown.
  Frames are from LytroZoom-O.
	  }

	\label{fig:compare2}
\end{figure*}

\subsection{Quantitative and Qualitative Comparisons}
In our comparative analysis, LF-DEST is benchmarked against a selection of advanced methods:
(1) DistgSSR~\cite{wang2022disentangling} and LFT~\cite{liang2022light}: these two methods are developed on the bicubic downsampling degradation, and achieve superior results on bicubic-based light field SR.
(2) SRMD~\cite{zhang2018learning}: a popular non-blind single image SR method developed on isotropic Gaussian blur and Gaussian noise degradation.
(3) DASR~\cite{wang2021unsupervised}: an advanced blind single image SR method developed on isotropic Gaussian blur and Gaussian noise degradation;
(4) BSRGAN~\cite{zhang2021designing} and Real-ESRGAN~\cite{wang2021real}: two real-world single image SR methods developed on the complex synthetic degradations.
(5) LF-DMnet: Pioneering the integration of blur and noise at multiple levels, LF-DMnet is the first deep network capable of adapting to different degradations in light field SR.
In addition to these advanced methods, we included the bicubic upsampling method for comprehensive benchmarking.

\noindent \textbf{Quantitative results on synthetic datasets.}
In Table~\ref{tab:quantitative}, we present the quantitative results of PSNR and SSIM obtained by various methods when applied to synthetic degradations characterized by diverse blur and noise levels.
LF-DEST consistently emerges as the top performer, showcasing its robustness and adaptability to a range of noise levels and kernel sizes.
Notably, when noise levels are set at 15 and 50, LF-DEST consistently achieves the highest PSNR and SSIM scores across the HCInew, HCIold, and STFgantry datasets. For instance, on the HCInew dataset with a kernel width of 1.5, LF-DEST outperforms LF-DMnet by a substantial margin at noise levels of 15 and 50, yielding impressive performance gains of 0.61/0.011 and 0.71/0.017, respectively.
LF-DEST does not consistently outperform other methods when noise levels are at their lowest (\textit{i.e.}, noise level is 0). This suggests that alternative approaches may be more effective in handling noise-free conditions.
However, LF-DEST maintains its competitive edge as noise levels increase, demonstrating its resilience in handling noisy data. Even at the highest noise level of 4.5, LF-DEST maintains strong performance with remarkable PSNR and SSIM scores.
In scenarios involving out-of-distribution degradations with a kernel width of 4.5, LF-DEST consistently delivers favorable results. This underscores LF-DEST's adaptability to diverse and complex degradation patterns, outperforming LF-DMnet.
In summary, LF-DEST's consistent superior performance, adaptability to various noise levels and kernel sizes, and its prowess in out-of-distribution scenarios make it a strong contender in light field SR.

\noindent \textbf{Qualitative results on synthetic datasets.}
In Figure~\ref{fig:compare1}, we present a visual comparison of the results produced by various methods across different blur kernel widths and noise levels.
Notably, the baseline methods exhibit certain shortcomings, notably the generation of artificial textures that are not true to the original content. Conversely, our proposed LF-DEST demonstrates a remarkable ability to recover genuine details from the input light fields, even when they are affected by substantial blurriness and noise.

\begin{table}[!t]
	\caption{Comparisons of the number of parameters (M), FLOPs (G) and running time (s) for 4$\times$SR. FLOPs and running time are calculated on an input light field with an angular resolution of $5\times5$ and a spatial resolution of $32\times32$.
	PSNR and SSIM are averaged over 9 degradations in Table~\ref{tab:quantitative}.
	}\label{tab:efficiency}
	\vspace{-3mm}
	\centering
	\tiny
	\setlength{\tabcolsep}{0.7mm}
	\resizebox{\linewidth}{!}{ 
	
	\begin{tabular}{|l|c|c|c|c|c|c|}
	\hline
	  &  \multirow{2}{*}{\#Param.}  & \multirow{2}{*}{FLOPs} & \multirow{2}{*}{Time} & \multicolumn{3}{c|}{PSNR$/$SSIM} \\
	  \cline{5-7}
		& & & & HCInew & HCIold & STFgantry \\
	\hline
	SRMD     & 1.50 & 39.76 & 0.070 & 27.18$/$0.838 & 31.16$/$0.913 & 25.78$/$0.840 \\
	DASR     & 5.80 & 82.03 & 0.051 & 26.74$/$0.831 & 29.47$/$0.896 & 24.92$/$0.829 \\
	BSRNet	  & 16.70 & 459.6 & 0.119 & 24.03$/$0.807 & 26.17$/$0.856 & 21.98$/$0.793 \\
	Real-ESRGAN  & 16.70 & 459.6 & 0.119 & 25.92$/$0.821 & 28.52$/$0.888 & 22.82$/$0.809 \\
	DistgSSR & 3.53 & 65.41 & 0.037 & 22.79$/$0.611 & 24.97$/$0.642 & 22.06$/$0.623 \\
	LFT & 1.11 & 29.45 & 0.070 & 22.92$/$0.612 & 25.23$/$0.647 & 22.14$/$0.623 \\
	LF-DMnet & 3.80 & 65.93 & 0.039 & {28.75}$/${0.869} & {33.69}$/${0.939} & {27.61}$/${0.885} \\
	LF-DEST w/o $\mathcal{N}_{DE}$ & 3.73 & 126.15 & 0.097 & 28.76$/$0.868 & 33.79$/$0.940 & 27.78$/$0.892 \\
	LF-DEST & 7.03 & 139.08 & 0.138 & 29.21$/$0.877 & 34.17$/$0.943 & 28.28$/$0.902 \\
	\hline
	\end{tabular}}
	\end{table}

\noindent \textbf{Computational efficiency.}
In Table~\ref{tab:efficiency}, we compare the efficiency of LF-DEST with baseline methods.
LF-DEST, with its comprehensive pipeline, including the degradation estimation part, exhibits higher computational complexity compared to LF-DMnet.
However, removing the degradation estimation part $\mathcal{N}_{DE}$ results in a more efficient LF-DEST variant.
This efficiency gain comes at a cost, with a noticeable performance drop in PSNR and SSIM results.
LF-DEST stands out in terms of reconstruction performance, albeit with increased computational costs.
Future improvements may include the incorporation of lightweight network designs and other optimizations to strike a balance between computational costs and performance.

\noindent \textbf{Qualitative results on real-world datasets.}
We evaluate the performance of various methods on a real-world degraded light field from the LytroZoom dataset, as depicted in Figure~\ref{fig:compare2}. 
We infer LF-DMNet with $N$ and $K$ values as indicated in Table~\ref{tab:quantitative} and select the optimal visual results for evaluation.
Notably, we observe that LF-DEST generates significantly improved results, indicating its strong ability to generalize effectively in real-world scenarios.

\begin{table}[!t]
\caption{Effectiveness of the degradation estimation in LF-DEST.
}\label{tab:degestimation}
\vspace{-3mm}
\centering
\tiny
\setlength{\tabcolsep}{1.8mm}
\resizebox{0.875\linewidth}{!}{ 

\begin{tabular}{|l|c|c|c|c|c|}
\hline
 \multirow{2}{*}{Method}  &  \multirow{2}{*}{Kernel}  & \multirow{2}{*}{Noise}  & \multicolumn{3}{c|}{PSNR$/$SSIM} \\
  \cline{4-6}
    & &  & HCInew & HCIold & STFgantry \\
\hline
(a) & \XSolidBrush & \XSolidBrush & 28.76$/$0.868 & 33.79$/$0.940 & 27.78$/$0.892  \\
(b) & \Checkmark & \XSolidBrush & 29.04$/$0.874 & 33.97$/$0.942 & 28.08$/$0.898 \\
(c) & \Checkmark & \XSolidBrush &29.05$/$0.874 & 34.10$/$0.943  & 28.15$/$0.899 \\
(d) & \XSolidBrush & \Checkmark &28.90$/$0.871 &  33.85$/$0.941 & 27.97$/$0.896 \\
(e) & \XSolidBrush & \Checkmark & 28.91$/$0.872 & 33.90$/$0.941 & 27.98$/$0.896 \\
(f) & \Checkmark & \Checkmark & 29.21$/$0.877 & 34.17$/$0.943 & 28.28$/$0.902 \\
\hline  
\end{tabular}
}
\end{table}

\begin{figure}[!t]
 \centering
 \begin{minipage}{0.9\linewidth}
\begin{minipage}{0.1910\linewidth}
\centerline{\includegraphics[width=1\linewidth]{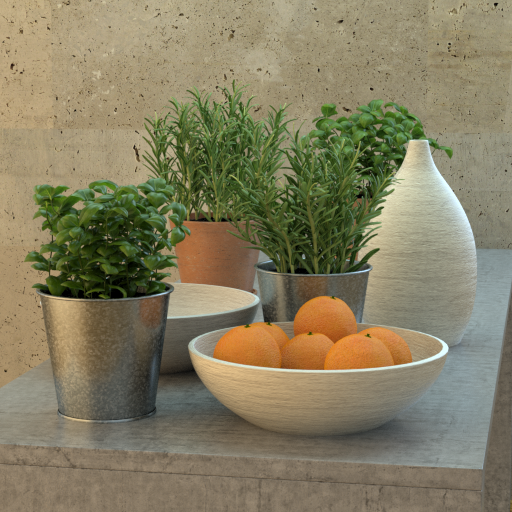}} \vfill \vspace{-0.15cm}
	\centerline{\tiny{HCI\_new\_0}}
	\end{minipage}
	\hfill
	\hspace{-0.200cm}
	\begin{minipage}{0.1910\linewidth}
\centerline{\includegraphics[width=1\linewidth]{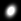}}\vfill \vspace{-0.15cm}
	\centerline{\tiny{KernelGAN}}
	\end{minipage}
 \hfill
	\hspace{-0.200cm}
	\begin{minipage}{0.1910\linewidth}
\centerline{\includegraphics[width=1\linewidth]{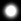}}\vfill \vspace{-0.15cm}
	\centerline{\tiny{$\mathcal{N}_{DE}$-Pretrained}}
	\end{minipage}
 \hfill
	\hspace{-0.200cm}
	\begin{minipage}{0.1910\linewidth}
\centerline{\includegraphics[width=1\linewidth]{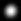}}\vfill \vspace{-0.15cm}
	\centerline{\tiny{$\mathcal{N}_{DE}$}}
	\end{minipage}
 \hfill
	\hspace{-0.200cm}
	\begin{minipage}{0.1910\linewidth}
\centerline{\includegraphics[width=1\linewidth]{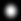}}\vfill \vspace{-0.15cm}
	\centerline{\tiny{GT}}
\end{minipage}
\end{minipage}
\vspace{-3mm}
 \caption{Visualizations of the estimated blur kernels generated by $\mathcal{N}_{DE}$ on the light field with an unknown blur kernel.
 }
\label{fig:kernels}
 \end{figure}

 \subsection{Model Analysis}
\begin{figure}[!t]
\vspace{-2mm} \centering
\begin{minipage}{0.9\linewidth}
	\begin{minipage}{0.1930\linewidth}
		\vspace{0.05cm}
\centerline{\includegraphics[width=1\linewidth]{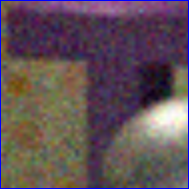}}\vfill 
\vspace{-0.15cm}
	\centerline{\tiny{Input}}
	\end{minipage}
 \hfill
	\hspace{-0.200cm}
	\begin{minipage}{0.1930\linewidth}
\centerline{\includegraphics[width=1\linewidth]{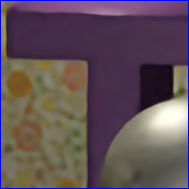}}\vfill 
\vspace{-0.15cm}
	\centerline{\tiny{KernelGAN}}
	\end{minipage}
 \hfill
	\hspace{-0.200cm}
	\begin{minipage}{0.1930\linewidth}
\centerline{\includegraphics[width=1\linewidth]{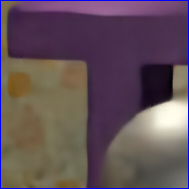}}\vfill 
\vspace{-0.15cm}
	\centerline{\tiny{CBDNet}}
	\end{minipage}
 \hfill
	\hspace{-0.200cm}
	\begin{minipage}{0.1930\linewidth}
\centerline{\includegraphics[width=1\linewidth]{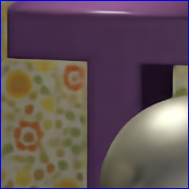}}\vfill 
\vspace{-0.15cm}
	\centerline{\tiny{LF-DEST}}
	\end{minipage}
 \hfill
	\hspace{-0.200cm}
	\begin{minipage}{0.1930\linewidth}
\centerline{\includegraphics[width=1\linewidth]{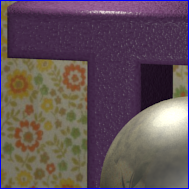}}\vfill 
\vspace{-0.15cm}
	\centerline{\tiny{GT}}
\end{minipage}
\end{minipage}
\vspace{-3mm}
 \caption{Visualizations of the final results generated from inaccurate estimated blur kernel (from KernelGAN) and inaccurate estimated noise map (from CBDNet).
 }
\label{fig:kernel_noise}
 \end{figure}

To analyze the designs in LF-DEST, we conduct the following experiments in terms of PSNR/SSIM.
We maintain unchanged parameters by utilizing residual blocks when removing different modules.
\noindent \textbf{Effectiveness of the degradation estimation.}
The degradation estimator $\mathcal{N}_{DE}$ simultaneously estimates blur kernels and noise maps from LR light fields, empowering LF-DEST with the ability to handle various degradation types and complex scenarios.
To comprehensively analyze the effectiveness of $\mathcal{N}_{DE}$, we design and execute a series of variants as follows:
(a) We directly remove $\mathcal{N}_{DE}$.
(b) We delete the branch responsible for generating noise maps.
(c) We remove the noise map branch and use the noise map estimated by CBDNet to generate $\bm{F}_{Deg}$.
(d) We remove the branch responsible for generating kernels.
(e) We remove the branch responsible for generating kernels and use KernelGAN to estimate the degradation kernel for generating $\bm{F}_{Deg}$.
Quantitative results are presented in Table~\ref{tab:degestimation}.
We have made several notable observations. Firstly, the significance of kernel estimation surpasses that of noise estimation. For instance, on HCINew, method (b) outperforms method (d) by 0.14/0.003 in terms of PSNR/SSIM.
Secondly, we emphasize the importance of degradation estimation in light field super-resolution. Even when inaccurate degradations are estimated by KernelGAN~\cite{bell2019blind} and CBDNet~\cite{guo2019toward}, they can still yield slight performance improvements. In contrast, our LF-DEST's degradation estimation method consistently produces the best results.
In Figure~\ref{fig:kernels}, we visualize the estimated blur kernels.
Compared to the blur kernel estimated by KernelGAN, the kernel estimated using $\mathcal{N}_{DE}$ is notably more accurate.
Moreover, in Figure~\ref{fig:kernel_noise}, we present the final reconstruction results using inaccurately estimated degradations by KernelGAN and CBDNet.
Our findings indicate that inaccurate kernel estimation leads to ringing artifacts and imprecise details, while inaccurate noise maps result in excessively smooth outputs.
These observations underscore the critical role of accurate degradation estimation in light field SR, with kernel estimation playing a particularly pivotal role.
Our LF-DEST excels in this regard, leading to superior performance and more faithful estimations of degradation components.

\begin{table}[!t]
\caption{Effectiveness of the designs in degradation estimator.
}\label{tab:degestimation2}
\vspace{-3mm}
\centering
\tiny
\setlength{\tabcolsep}{1.8mm}
\resizebox{0.875\linewidth}{!}{ 

\begin{tabular}{|l|c|c|c|c|c|}
\hline
 \multirow{2}{*}{Method}  &  \multirow{2}{*}{S2C}  & \multirow{2}{*}{C2S}  & \multicolumn{3}{c|}{PSNR$/$SSIM} \\
  \cline{4-6}
    & &  & HCInew & HCIold & STFgantry \\
\hline
(a) & \XSolidBrush & \XSolidBrush & 29.11$/$0.875 & 34.11$/$0.943 & 28.22$/$0.901 \\
(b) & \Checkmark & \XSolidBrush & 29.17$/$0.877 & 34.10$/$0.943 & 28.26$/$0.901 \\
(c) & \XSolidBrush & \Checkmark & 29.13$/$0.876 & 34.16$/$0.943 & 28.24$/$0.902 \\
(d) & \Checkmark & \Checkmark & 29.21$/$0.877 & 34.17$/$0.943 & 28.28$/$0.902 \\
\hline  
\end{tabular}
}
\end{table}

\noindent \textbf{Effectiveness of the components in the degradation estimator.}
The side-to-center fusion block and the center-to-side fusion block are pivotal components within the degradation estimator.
Both blocks are designed to harness the complete potential of angular and spatial correlations for accurate degradation estimation. 
We design and execute a series of variants as follows:
(a) We directly remove both blocks and replace them with residual blocks, maintaining parameter stability.
(b) We remove the center-to-side fusion block.
(c) We remove the side-to-center fusion block.
Based on the results in Table~\ref{tab:degestimation2}, (d) achieves the best performance across all three datasets, with a PSNR of 29.21 and SSIM of 0.877.
This suggests that comprehensive information utilization between side-view and center-view, as implemented in LF-DEST, is beneficial for improving reconstruction performance.
(b) and (c) outperform (a), indicating that using a single block (either from side-view to center-view or from center-view to side-view) can enhance performance, although not as effectively as bidirectional utilization.

\begin{table}[!t]
\caption{Effectiveness of the MSF module in LF-DEST.
}\label{tab:msf}
\vspace{-3mm}
\centering
\tiny
\setlength{\tabcolsep}{1.8mm}
\resizebox{0.875\linewidth}{!}{ 

\begin{tabular}{|l|c|c|c|c|c|}
\hline
 \multirow{2}{*}{Method}  &  \multirow{2}{*}{CrA}  & \multirow{2}{*}{AW}  & \multicolumn{3}{c|}{PSNR$/$SSIM} \\
  \cline{4-6}
    & &  & HCInew & HCIold & STFgantry \\
\hline
(a) & \XSolidBrush & \XSolidBrush & 29.11$/$0.875 & 33.97$/$0.942 & 28.24$/$0.901 \\
(b) & \Checkmark & \XSolidBrush & 29.13$/$0.875 & 34.11$/$0.943 & 28.26$/$0.902 \\
(c) & \XSolidBrush & \Checkmark & 29.13$/$0.876 & 34.10$/$0.943 & 28.23$/$0.902 \\
(d) & \Checkmark & \Checkmark & 29.21$/$0.877 & 34.17$/$0.943 & 28.28$/$0.902 \\
\hline  
\end{tabular}
}
\end{table}

\noindent \textbf{Effectiveness of the MSF module.}
The MSF module's ability to adaptively fuse degradation representations and image features is the essence of blind light field SR.
We design following variants:
(a) We directly remove the CrA blocks and the AW layers and replace them with residual blocks, maintaining parameter stability.
(b) We remove the AW layers.
(c) We remove the CrA blocks.
Based on the results in Table~\ref{tab:msf}, (d) achieves the best performance across all three datasets.
This demonstrates the significance of both designs in enhancing the performance of LF-DEST across various datasets.

\begin{figure}[!t]
 \vspace{-2mm}\centering
 \begin{minipage}{0.7\linewidth}
	\begin{minipage}{0.2430\linewidth}
\vspace{0.05cm}
\centerline{\includegraphics[width=1\linewidth]{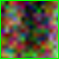}}\vfill 
\centerline{\includegraphics[width=1\linewidth]{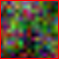}}\vfill 
\vspace{-0.15cm}
	\centerline{\tiny{Input}}
	\end{minipage}
 \hfill
	\hspace{-0.200cm}
	\begin{minipage}{0.2430\linewidth}
\centerline{\includegraphics[width=1\linewidth]{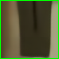}}\vfill 
\centerline{\includegraphics[width=1\linewidth]{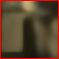}}\vfill 
\vspace{-0.15cm}
	\centerline{\tiny{LF-DMnet}}
	\end{minipage}
 \hfill
	\hspace{-0.200cm}
	\begin{minipage}{0.2430\linewidth}
\centerline{\includegraphics[width=1\linewidth]{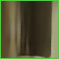}}\vfill 
\centerline{\includegraphics[width=1\linewidth]{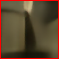}}\vfill 
\vspace{-0.15cm}
	\centerline{\tiny{LF-DEST}}
	\end{minipage}
 \hfill
	\hspace{-0.200cm}
	\begin{minipage}{0.2430\linewidth}
\centerline{\includegraphics[width=1\linewidth]{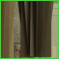}}\vfill 
\centerline{\includegraphics[width=1\linewidth]{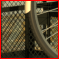}}\vfill 
\vspace{-0.15cm}
	\centerline{\tiny{GT}}
	\end{minipage}
 \end{minipage}
 \vspace{-3mm}
 \caption{Fail cases. 
    LF-DEST performs poor on severely degraded scenes with fine details (kernel width=$4.5$ and noise level=$15$).
 }
\label{fig:limitation}
 \end{figure}

\subsection{Limitations and Discussions}
While our proposed LF-DEST has shown promising performance, it does have limitations when faced with challenging scenarios.
For instance, LF-DEST struggles to recover precise details and textures, particularly in cases of severely degraded light field images (see Figure~\ref{fig:limitation}).
Additionally, our LF-DEST model primarily focuses on modeling the impact of blur and noise on light field SR and does not explicitly account for other degradations commonly encountered in real-world scenarios, such as rain or haze. These unmodeled degradations remain an avenue for future improvements.
In our future work, we plan to address these challenges and incorporate more complex degradations into the LF-DEST framework, making it more robust and applicable to a wider range of scenarios.
We also aim to enhance the efficiency to reduce computational costs, ensuring that it remains practical for real-time or large-scale applications.

Current light field SR methods often lack accurate light path modeling, representing a \textit{common limitation}.
As an extension of these methods, our exploration of light field SR within real and complex scenes serves as an inspiring piece that could pave the way for future work.
An additional noteworthy constraint of LF-DEST lies in its reliance on the degradation model referenced in~\cite{wang2022learning}. While this model has demonstrated utility in certain intricate scenarios, it encounters several challenges. 
Firstly, the issue pertains to the degradation itself: within a real light field camera setup, the blur kernel may exhibit angular variation (and spatially if lens aberrations are considered). 
Evaluating the impact of highly simplified simulations proves arduous from the presented findings. Secondly, degradation encompasses factors extending beyond mere blur and noise. 
Parameters such as main lens focal length, inter-sensor distance, lenslet-sensor separation, photosensor pitch size, inter-lenslet distance, and lenslet aperture size contribute to this complexity. 
Future efforts integrating these parameters could yield more precise and authentic degradation modeling, aligning better with light field camera characteristics. 
This, in turn, would refine simulation data and fortify the generalization capacity of the new algorithms.

\section{Conclusion}
In this paper, we introduce LF-DEST, an effective light field SR method designed to tackle a range of degradation types without relying on fixed degradation models like bicubic downsampling.
LF-DEST consists of two primary components: degradation estimation and light field restoration.
The former concurrently estimates blur kernels and noise maps from LR input light fields, while the latter generates super-resolved light field data based on the estimated degradation representations.
Our comprehensive experiments, conducted on synthetic and real-world datasets, validate that LF-DEST outperforms existing methods across a variety of light field SR degradation scenarios.

{\small
\normalem
\bibliographystyle{ieee}
\bibliography{bib}

\begin{thebibliography}{10}\itemsep=-1pt

\bibitem{alain2018light}
Martin Alain and Aljosa Smolic.
\newblock Light field super-resolution via lfbm5d sparse coding.
\newblock In {\em ICIP}, 2018.

\bibitem{bell2019blind}
Sefi Bell-Kligler, Assaf Shocher, and Michal Irani.
\newblock Blind super-resolution kernel estimation using an internal-gan.
\newblock {\em NeurlPS}, 32, 2019.

\bibitem{bergen1991plenoptic}
James~R Bergen and Edward~H Adelson.
\newblock The plenoptic function and the elements of early vision.
\newblock {\em Computational models of visual processing}, 1:8, 1991.

\bibitem{cheng2022spatial}
Zhen Cheng, Yutong Liu, and Zhiwei Xiong.
\newblock Spatial-angular versatile convolution for light field reconstruction.
\newblock {\em IEEE Transactions on Computational Imaging}, 8:1131--1144, 2022.

\bibitem{cheng2021light}
Zhen Cheng, Zhiwei Xiong, Chang Chen, Dong Liu, and Zheng-Jun Zha.
\newblock Light field super-resolution with zero-shot learning.
\newblock In {\em CVPR}, 2021.

\bibitem{cheng2019light}
Zhen Cheng, Zhiwei Xiong, and Dong Liu.
\newblock Light field super-resolution by jointly exploiting internal and
  external similarities.
\newblock {\em IEEE Transactions on Circuits and Systems for Video Technology},
  30(8):2604--2616, 2019.

\bibitem{cong2023lfdet}
Ruixuan Cong, Hao Sheng, Da Yang, Zhenglong Cui, and Rongshan Chen.
\newblock Exploiting spatial and angular correlations with deep efficient
  transformers for light field image super-resolution.
\newblock {\em IEEE Transactions on Multimedia}, 2023.

\bibitem{dong2014learning}
Chao Dong, Chen~Change Loy, Kaiming He, and Xiaoou Tang.
\newblock Learning a deep convolutional network for image super-resolution.
\newblock In {\em ECCV}, 2014.

\bibitem{gu2019blind}
Jinjin Gu, Hannan Lu, Wangmeng Zuo, and Chao Dong.
\newblock Blind super-resolution with iterative kernel correction.
\newblock In {\em CVPR}, 2019.

\bibitem{guo2019toward}
Shi Guo, Zifei Yan, Kai Zhang, Wangmeng Zuo, and Lei Zhang.
\newblock Toward convolutional blind denoising of real photographs.
\newblock In {\em Proceedings of the IEEE/CVF conference on computer vision and
  pattern recognition}, pages 1712--1722, 2019.

\bibitem{HCInew}
Katrin Honauer, Ole Johannsen, Daniel Kondermann, and Bastian Goldluecke.
\newblock A dataset and evaluation methodology for depth estimation on 4d light
  fields.
\newblock In {\em ACCV}, 2016.

\bibitem{jin2020light}
Jing Jin, Junhui Hou, Jie Chen, and Sam Kwong.
\newblock Light field spatial super-resolution via deep combinatorial geometry
  embedding and structural consistency regularization.
\newblock In {\em CVPR}, 2020.

\bibitem{levoy1996light}
Marc Levoy and Pat Hanrahan.
\newblock Light field rendering.
\newblock In {\em Proceedings of the 23rd annual conference on Computer
  graphics and interactive techniques}, pages 31--42, 1996.

\bibitem{liang2015light}
Chia-Kai Liang and Ravi Ramamoorthi.
\newblock A light transport framework for lenslet light field cameras.
\newblock {\em ACM Transactions on Graphics (TOG)}, 34(2):1--19, 2015.

\bibitem{liang2021flow}
Jingyun Liang, Kai Zhang, Shuhang Gu, Luc Van~Gool, and Radu Timofte.
\newblock Flow-based kernel prior with application to blind super-resolution.
\newblock In {\em Proceedings of the IEEE/CVF Conference on Computer Vision and
  Pattern Recognition}, pages 10601--10610, 2021.

\bibitem{liang2022light}
Zhengyu Liang, Yingqian Wang, Longguang Wang, Jungang Yang, and Shilin Zhou.
\newblock Light field image super-resolution with transformers.
\newblock {\em IEEE Signal Processing Letters}, 29:563--567, 2022.

\bibitem{liang2023learning}
Zhengyu Liang, Yingqian Wang, Longguang Wang, Jungang Yang, Shilin Zhou, and
  Yulan Guo.
\newblock Learning non-local spatial-angular correlation for light field image
  super-resolution.
\newblock {\em arXiv preprint arXiv:2302.08058}, 2023.

\bibitem{liu2021intra}
Gaosheng Liu, Huanjing Yue, Jiamin Wu, and Jingyu Yang.
\newblock Intra-inter view interaction network for light field image
  super-resolution.
\newblock {\em IEEE Transactions on Multimedia}, 2021.

\bibitem{luo2020unfolding}
Zhengxiong Luo, Yan Huang, Shang Li, Liang Wang, and Tieniu Tan.
\newblock Unfolding the alternating optimization for blind super resolution.
\newblock 2020.

\bibitem{luo2021endtoend}
Zhengxiong Luo, Yan Huang, Shang Li, Liang Wang, and Tieniu Tan.
\newblock End-to-end alternating optimization for blind super resolution.
\newblock {\em arXiv preprint arXiv:2105.06878}, 2021.

\bibitem{luo2021end}
Zhengxiong Luo, Yan Huang, Shang Li, Liang Wang, and Tieniu Tan.
\newblock End-to-end alternating optimization for blind super resolution.
\newblock {\em arXiv preprint arXiv:2105.06878}, 2021.

\bibitem{meng2019high}
Nan Meng, Hayden Kwok-Hay So, Xing Sun, and Edmund Lam.
\newblock High-dimensional dense residual convolutional neural network for
  light field reconstruction.
\newblock {\em IEEE transactions on pattern analysis and machine intelligence},
  2019.

\bibitem{ng2005light}
Ren Ng, Marc Levoy, Mathieu Br{\'e}dif, Gene Duval, Mark Horowitz, and Pat
  Hanrahan.
\newblock {\em Light field photography with a hand-held plenoptic camera}.
\newblock PhD thesis, Stanford University, 2005.

\bibitem{pan2021deep}
Jinshan Pan, Haoran Bai, Jiangxin Dong, Jiawei Zhang, and Jinhui Tang.
\newblock Deep blind video super-resolution.
\newblock In {\em ICCV}, 2021.

\bibitem{rossi2017graph}
Mattia Rossi and Pascal Frossard.
\newblock Graph-based light field super-resolution.
\newblock In {\em 2017 IEEE 19th International Workshop on Multimedia Signal
  Processing (MMSP)}, 2017.

\bibitem{shocher2018zero}
Assaf Shocher, Nadav Cohen, and Michal Irani.
\newblock “zero-shot” super-resolution using deep internal learning.
\newblock In {\em Proceedings of the IEEE conference on computer vision and
  pattern recognition}, pages 3118--3126, 2018.

\bibitem{tao2021spectrum}
Guangpin Tao, Xiaozhong Ji, Wenzhuo Wang, Shuo Chen, Chuming Lin, Yun Cao, Tong
  Lu, Donghao Luo, and Ying Tai.
\newblock Spectrum-to-kernel translation for accurate blind image
  super-resolution.
\newblock In {\em Thirty-Fifth Conference on Neural Information Processing
  Systems}, 2021.

\bibitem{STFgantry}
Vaibhav Vaish and Andrew Adams.
\newblock The (new) stanford light field archive.
\newblock {\em Computer Graphics Laboratory, Stanford University}, 6(7), 2008.

\bibitem{van2023end}
Vinh Van~Duong, Thuc~Nguyen Huu, Jonghoon Yim, and Byeungwoo Jeon.
\newblock End-to-end learned light field image rescaling using joint
  spatial-angular and epipolar information.
\newblock In {\em 2023 IEEE International Conference on Image Processing
  (ICIP)}, pages 1935--1939. IEEE, 2023.

\bibitem{van2023light}
Vinh Van~Duong, Thuc~Nguyen Huu, Jonghoon Yim, and Byeungwoo Jeon.
\newblock Light field image super-resolution network via joint spatial-angular
  and epipolar information.
\newblock {\em IEEE Transactions on Computational Imaging}, 9:350--366, 2023.

\bibitem{wang2021unsupervised}
Longguang Wang, Yingqian Wang, Xiaoyu Dong, Qingyu Xu, Jungang Yang, Wei An,
  and Yulan Guo.
\newblock Unsupervised degradation representation learning for blind
  super-resolution.
\newblock In {\em CVPR}, 2021.

\bibitem{wang2022detail}
Shunzhou Wang, Tianfei Zhou, Yao Lu, and Huijun Di.
\newblock Detail preserving transformer for light field image super-resolution.
\newblock In {\em AAAI}, 2022.

\bibitem{wang2015occlusion}
Ting-Chun Wang, Alexei~A Efros, and Ravi Ramamoorthi.
\newblock Occlusion-aware depth estimation using light-field cameras.
\newblock In {\em ICCV}, 2015.

\bibitem{wang2021real}
Xintao Wang, Liangbin Xie, Chao Dong, and Ying Shan.
\newblock Real-esrgan: Training real-world blind super-resolution with pure
  synthetic data.
\newblock In {\em ICCV}, 2021.

\bibitem{wang2022learning}
Yingqian Wang, Zhengyu Liang, Longguang Wang, Jungang Yang, Wei An, and Yulan
  Guo.
\newblock Real-world light field image super-resolution via degradation
  modulation.
\newblock {\em arXiv preprint arXiv:2206.06214}, 2022.

\bibitem{wang2018lfnet}
Yunlong Wang, Fei Liu, Kunbo Zhang, Guangqi Hou, Zhenan Sun, and Tieniu Tan.
\newblock Lfnet: A novel bidirectional recurrent convolutional neural network
  for light-field image super-resolution.
\newblock {\em IEEE Transactions on Image Processing}, 27(9):4274--4286, 2018.

\bibitem{wang2022disentangling}
Yingqian Wang, Longguang Wang, Gaochang Wu, Jungang Yang, Wei An, Jingyi Yu,
  and Yulan Guo.
\newblock Disentangling light fields for super-resolution and disparity
  estimation.
\newblock {\em IEEE Transactions on Pattern Analysis and Machine Intelligence},
  2022.

\bibitem{wang2020spatial}
Yingqian Wang, Longguang Wang, Jungang Yang, Wei An, Jingyi Yu, and Yulan Guo.
\newblock Spatial-angular interaction for light field image super-resolution.
\newblock In {\em ECCV}, 2020.

\bibitem{wang2020light}
Yingqian Wang, Jungang Yang, Longguang Wang, Xinyi Ying, Tianhao Wu, Wei An,
  and Yulan Guo.
\newblock Light field image super-resolution using deformable convolution.
\newblock {\em IEEE Transactions on Image Processing}, 30:1057--1071, 2020.

\bibitem{HCIold}
Sven Wanner, Stephan Meister, and Bastian Goldluecke.
\newblock Datasets and benchmarks for densely sampled 4d light fields.
\newblock In {\em Vision, Modelling and Visualization (VMV)}, volume~13, pages
  225--226. Citeseer, 2013.

\bibitem{Xiao_2023_toward}
Zeyu Xiao, Ruisheng Gao, Yutong Liu, Yueyi Zhang, and Zhiwei Xiong.
\newblock Toward real-world light field super-resolution.
\newblock In {\em CVPRW}, 2023.

\bibitem{Xiao_2023_cutmib}
Zeyu Xiao, Yutong Liu, Ruisheng Gao, and Zhiwei Xiong.
\newblock Cutmib: Boosting light field super-resolution via multi-view image
  blending.
\newblock In {\em CVPR}, 2023.

\bibitem{yeung2018light}
Henry Wing~Fung Yeung, Junhui Hou, Xiaoming Chen, Jie Chen, Zhibo Chen, and
  Yuk~Ying Chung.
\newblock Light field spatial super-resolution using deep efficient
  spatial-angular separable convolution.
\newblock {\em IEEE Transactions on Image Processing}, 28(5):2319--2330, 2018.

\bibitem{yoon2017light}
Youngjin Yoon, Hae-Gon Jeon, Donggeun Yoo, Joon-Young Lee, and In~So Kweon.
\newblock Light-field image super-resolution using convolutional neural
  network.
\newblock {\em IEEE Signal Processing Letters}, 24(6):848--852, 2017.

\bibitem{yuan2018light}
Yan Yuan, Ziqi Cao, and Lijuan Su.
\newblock Light-field image superresolution using a combined deep cnn based on
  epi.
\newblock {\em IEEE Signal Processing Letters}, 25(9):1359--1363, 2018.

\bibitem{zhang2021designing}
Kai Zhang, Jingyun Liang, Luc Van~Gool, and Radu Timofte.
\newblock Designing a practical degradation model for deep blind image
  super-resolution.
\newblock In {\em ICCV}, 2021.

\bibitem{zhang2018learning}
Kai Zhang, Wangmeng Zuo, and Lei Zhang.
\newblock Learning a single convolutional super-resolution network for multiple
  degradations.
\newblock In {\em CVPR}, 2018.

\bibitem{zhang2019residual}
Shuo Zhang, Youfang Lin, and Hao Sheng.
\newblock Residual networks for light field image super-resolution.
\newblock In {\em CVPR}, 2019.

\bibitem{zhang2021removing}
Shuo Zhang, Zeqi Shen, and Youfang Lin.
\newblock Removing foreground occlusions in light field using micro-lens
  dynamic filter.
\newblock In {\em IJCAI}, 2021.

\bibitem{zhang2016robust}
Shuo Zhang, Hao Sheng, Chao Li, Jun Zhang, and Zhang Xiong.
\newblock Robust depth estimation for light field via spinning parallelogram
  operator.
\newblock {\em Computer Vision and Image Understanding}, 145:148--159, 2016.

\bibitem{zhang2018residual}
Yulun Zhang, Yapeng Tian, Yu Kong, Bineng Zhong, and Yun Fu.
\newblock Residual dense network for image super-resolution.
\newblock In {\em CVPR}, 2018.

\end{thebibliography}
}

\end{document}